\documentclass[journal]{IEEEtran}

\usepackage{helvet}
\usepackage{courier}
\usepackage{bm}

\usepackage{graphicx}
\usepackage{subfigure}
\usepackage{multirow}%
\usepackage{amsmath, amssymb}
\usepackage{color}
\usepackage{mathrsfs}
\usepackage{amsthm}
\usepackage{epstopdf}
\usepackage{wrapfig}
\usepackage{picinpar}
\usepackage{url}
\usepackage{algorithm}
\usepackage{algorithmic}

\def\diag{\mbox{diag}}
\def\support{\mbox{support}}

\def\R{{\mathbb R}}

\def\0{{\bf 0}}
\def\1{{\bf 1}}

\def\bA {{\bf A}}
\def\mA {\mathcal{A}}

\def\Q{{\bf Q}}

\def\I{{\bf I}}

\def\s{{\bf s}}
\def\bg{{\bf g}}
\def\bR{{\bf R}}
\def\bX{{\bf X}}
\def\bB{{\bf B}}

\def\bfeta{\mbox{{\boldmath $\eta$}}}

\def\mJ{{\mathcal J}}

\def\mT{{\mathcal T}}

\def\mN{{\mathcal N}}
\def\mI{{\mathcal I}}
\def\mJ{{\mathcal J}}

\def\vv{{\bf v}}

\def\be{{\bf e}}

\def\x{{\bf x}}
\def\vv{{\bf v}}

\def\z{{\bf z}}

\def\bu{{\bf u}}

\def \dd{\mbox{{\boldmath $\tau$}}}
 \def \bfeta {\mbox{{\boldmath $\tau$}}}

\def\bPsi{\mbox{{\boldmath $\Psi$}}}
\def\ba{\mbox{{\boldmath $\alpha$}}}

\def\bb{\mbox{{\boldmath $\beta$}}}

\def\bxi{\mbox{{\boldmath $\xi$}}}

\def\db{{\bf b}}

\def\bA{{\bf A}}
\def\bB{{\bf B}}
\def\x{{\bf x}}

\def\RR{{\mathbb R}}

\def\revise{\textcolor{black}}
\def\citep{\cite}

\newtheorem{lemma}{Lemma}
\newtheorem{remark}{Remark}
\ifCLASSOPTIONcompsoc
\else
\fi

\ifCLASSINFOpdf
\else
\fi

\begin{document}
%
\title{Matching Pursuit LASSO Part II: Applications and Sparse Recovery over Batch Signals
}
%
%
%
%

\author{Mingkui~Tan,~
        Ivor W. Tsang,~
        and Li Wang
\IEEEcompsocitemizethanks{\IEEEcompsocthanksitem Mingkui Tan is with the School of Computer Science, the University of Adelaide, Australia. e-mail: mingkui.tan@adelaide.edu.au.  \protect\\
\IEEEcompsocthanksitem Ivor W. Tsang is with the Centre for Quantum Computation $\&$ Intelligent Systems (QCIS), at the University of Technology, Sydney (UTS), Australia. e-mail: Ivor.Tsang@uts.edu.au. \protect\\
\IEEEcompsocthanksitem Li Wang is with the Institute for Computational and Experimental Research in Mathematics (ICERM), Brown University, USA. e-mail: liwangucsd@gmail.com.}
\thanks{}}

\IEEEcompsoctitleabstractindextext{%
\begin{abstract}
In Part I~\cite{TanPMLPart1}, a Matching Pursuit LASSO
({MPL}) algorithm has been presented for solving large-scale sparse
recovery (SR) problems. In this paper, we  present a subspace
search to further improve the performance of MPL, and then continue to
address another major challenge of SR -- batch SR with many signals,
a consideration which is absent from most of previous $\ell_1$-norm methods. As a result, a batch-mode {MPL} is developed to
vastly speed up sparse recovery of many signals simultaneously.
Comprehensive numerical experiments on compressive sensing and face
recognition tasks demonstrate the superior performance of MPL and BMPL over
other methods considered in this paper, in terms of sparse recovery
ability and efficiency. In particular, BMPL is up to 400 times faster than existing $\ell_1$-norm methods considered to be state-of-the-art.
\end{abstract}
\begin{keywords}
Batch mode LASSO, sparse recovery, big dictionary, compressive
sensing, face recognition.
\end{keywords}}


\maketitle

\IEEEdisplaynotcompsoctitleabstractindextext

%
\IEEEpeerreviewmaketitle

\section{Introduction}\label{Introduction}
%
With the fast development of compressive sensing
theory~\cite{Candes2005}, sparse recovery (SR) has gained increased attention recently in the signal processing
community~\cite{Candes2005,David2006,duarte2011structured,do2012fast}.
It has also become a fundamental element of many other research
areas, such as image processing, computer vision, data mining and
machine
learning~\cite{Mairal2008Color,Wright2009,Elhamifar2009,Amir2011,ICML2011Coates,Peleg2012}.

Formally, SR seeks to recover an unknown $k$-sparse signal $\x \in
\R^m$ from its nonadaptive linear  measurement $\db  = \bA \x + \be
\in \R^n$, where $\bA \in \RR^{n\times m} (n \ll m)$ denotes the
dictionary, $\be \in \RR^n$ represents the  noise, and each column
vector of $\bA$ is referred to as an atom. To recover
$\x$ from $\db$, one need to solve an $\ell_0$-norm minimization
problem:
\begin{eqnarray}\label{eq:l0}
\min_\x ~~ \|\x\|_0, ~~ \text{s.t.} ~~  \db = \bA \x,
\end{eqnarray}
where $\|\cdot\|_{0}$ denotes the $\ell_{0}$-norm of a vector.
Problem (\ref{eq:l0}) is
NP-complete~\cite{davis1997adaptive,Candes2005,Ge2011}, and many researchers propose to solve its $\ell_1$-convex
relaxations instead~\cite{Efron2004, Honglak2006,David2006}, such as the following LASSO problem~\cite{kim2007interior,Figueiredo2007,linxiao,Xiao2013PGH_J}:
\begin{eqnarray}\label{eq:primall_2}
\min_\x ~~ \lambda \|\x\|_1 + \frac{1}{2}\|\db -\bA \x\|^2,
\end{eqnarray}
 where $\lambda$ is a regularization
parameter. Regarding problem (\ref{eq:primall_2}),  many methods have been
proposed over the last decade, such as the least-angle
regression (LARS)~\cite{efron2004least}, gradient projection for
sparse reconstruction (GPSR)~\cite{Figueiredo2007},  projected
gradient (PG)~\cite{Nesterov2007}, fast iterative
shrinkage-threshold algorithm ({FISTA})~\cite{Beck2009}, coordinate
descent methods~\cite{yun2011coordinate}, proximal gradient homotopy
(PGH) method~\cite{linxiao,Xiao2013PGH_J} and so on.  Interested readers can refer to Part I and the references
therein~\cite{TanPMLPart1} for a more comprehensive
review.

Existing $\ell_1$-norm methods, however, suffer from
high computational complexity for large-scale
SR problems. More critically, for problems like
\emph{batch SR}~\cite{Rubinstein2008}, in which many
signals need to be sparsely recovered simultaneously, the
computations will be even more expensive. Here, the
{batch SR} problem is carried out to solve the following optimization
problem:
\begin{eqnarray}\label{eq:batch}
\min_\bX ~~  \|\bB- \bA \bX\|_F^2 + \lambda \sum_{i=1}^p \|\x_i\|_1,
\end{eqnarray}
where $\bB = [\db_1, ..., \db_p] \in \R^{n\times p}$ records the
measurements of $p$ signals and $\|\cdot\|_F$ denotes the $F$-norm
of a matrix.  The batch SR problem plays an important role in many
applications, such as  face recognition~\cite{Wright2009,Yang2010ICIP}, compressive sensing~\cite{Baraniuk2007,romberg2008imaging},
dictionary learning~\cite{Aharon2006,lee2006efficient} and so on.

\subsection{Batch SR in Face Recognition}

Face recognition by {SR} has achieved promising performance recently~\cite{Wright2009,Yang2010ICIP,Gao2013,deng2012extended,zhuang2013single}. The
basic assumption is that, any testing image lies in
a subspace spanned by the training images of a
person~\cite{Wright2009,shi2011,Yang2010ICIP}, thus it can be
sparsely represented by the training images. Here, the training
images are formed as a dictionary $\bA \in \R^{n\times m}$, where $n$ denotes the number of pixels or features of a face image, and $m$ denotes the number of
training images. The core task
of {SR} based face recognition is to find a sparse representation
of a testing image $\db$ over $\bA$. However, directly solving problem (\ref{eq:primall_2}) is
computationally expensive especially when $n$ is
very large~\cite{Wright2009,shi2011,Yang2010ICIP}. Some researchers
propose to reduce the computational cost by dimension reduction methods, such as random projections~\cite{Wright2009}.
However, the recognition rates may be affected due to the dimension reduction~\cite{shi2011,Yang2010ICIP,ICML2011Coates}.

In practice, it is often required to recognize many
face images simultaneously in real-time, which is very challenging
for SR based methods~\cite{shi2011,Lei2011}. To address this,
the authors in \cite{shi2011} suggest directly solving
$\min_{\x}~\frac{1}{2}\|\db- \bA\x\|^2$, which is denoted by {L2}; while the
authors in \cite{Lei2011} argue that solving a least square problem
$\min_{\x}~\frac{1}{2}||\db-\bA\x||^2 + \frac{\lambda}{2}||\x||^2$, which is denoted by {L2-L2}, can achieve more
stable performance.
For the {L2} method, the optimal solution is {$\x^* = \bR^{+}\Q^{\top}\db$},
where $\bA = \Q\bR$ denotes the QR decomposition of $\bA$, and
$\bR^{+}$ denotes the pseudo inverse. For the {L2-L2} method, the
optimal solution is $\x^* =
(\bA^{\top}\bA+\lambda\I)^{-1}\bA^{\top}\db$. Therefore, fast predictions can be achieved via simple matrix-vector products by pre-computing $\bR^{-1}\Q^{\top}$ and $(\bA^{\top}\bA+\lambda\I)^{-1}$ off-line. However, since the solutions of the two methods are not sparse, the recognition performance may be degraded.

\subsection{Batch SR in Compressive Sensing}
Sparse recovery is a core element of the recently developed
compressive sensing theory on signal acquisition~\cite{Candes2005}. In compressive
sensing, a signal is allowed to be captured at a rate significantly
lower than the Nyquist rate, if it is
compressible or can be sparsely decomposed under a basis $\bPsi =
[\Psi_1, ..., \Psi_m]\in \R^{m\times
m}$~\cite{David2006,Baraniuk2007}.
To recover the original signal, we need to solve a
sparse recovery
problem~\cite{Baraniuk2007,romberg2008imaging}, which might be very expensive.
Moreover, in real-world sensing tasks, such as imaging and video
sensing~\cite{romberg2008imaging,Huang2013}, it is often necessary to
sense a large number of signals simultaneously in real-time.
Therefore, it is critical to efficiently address the
large-scale batch SR problem  in compressive sensing.

\subsection{Batch SR in Dictionary Learning}
Dictionary learning, which aims to find a good
dictionary based on a set of training signals, has recently  become
increasingly important in many areas, such as signal processing,
computer vision and machine
learning~\cite{lee2006efficient,Rubinstein2008,mairal2009online,
mairal2010online,Rubinstein2013}. To learn a good dictionary,
 many training examples (or signals) are usually required to be
sparsely represented at the same time, leading to an intolerable cost
for dictionary learning. The large-scale batch SR problem
therefore is a core step in dictionary learning~\cite{lee2006efficient,mairal2009online}.

\subsection{Main Contributions}

In Part I of this paper, we has presented a matching pursuit LASSO (MPL)
algorithm in relation to the computational issues of LASSO over big dictionaries.
In this paper, we first present a subspace
search to further improve the performance of MPL, and then continue to address the
computational bottleneck created by the batch SR problem. The main contributions of this paper are
summarized as follows:
\begin{itemize}
\item
A subspace exploratory matching is  proposed to  improve the
performance of MPL. This new
matching pursuit scheme takes less than 50 seconds to recover a $600$-sparse signal over a dictionary of
one million atoms.
\item
A batch mode {MPL} (BMPL), which is absent in many $\ell_1$-norm methods, is presented to address large-scale batch SR
problems.
\item
We apply {BMPL} to face recognition tasks on two well-known face
{databases}, namely \emph{{Extended YaleB}} and \emph{AR}
{databases}. Comprehensive experiments show that {BMPL} achieves
comparable or better recognition rates than baselines with comparable time complexity. Importantly, {BMPL} is up to 400 times faster than existing $\ell_1$-norm methods considered to be state-of-the-art.
\end{itemize}

The rest of this paper is organized as follows. In
Section~\ref{sec:MPL},  we briefly review the MPL algorithm and then
propose an improved MPL algorithm with subspace exploratory
matching. In Section~\ref{sec:batch}, we describe the
{batch mode MPL} method.  Numerical
experiments and real-world applications are presented in
Sections~\ref{sec:num_exp} and \ref{exp:batch}, respectively.
Conclusive remarks are given in Section \ref{sec:con}.

\section{Matching Pursuit for LASSO}\label{sec:MPL}

Throughout the paper, we denote the transpose of a vector/matrix by
the superscript $^{\top}$, $\0$ as a zero vector and $\diag(\vv)$ as
a diagonal matrix with diagonal entries equal to $\vv$. In addition,
let $\|\vv\|_p$ and $\|\vv\|$ denote the $\ell_p$-norm and
$\ell_2$-norm of a vector $\vv$, respectively. For a function
$f(\x)$, let $\nabla f(\x)$ and $\partial f(\x)$ be the
gradient and subgradient of $f(\x)$ at $\x$, respectively. For a
sparse vector $\x$, let the calligraphic letter $\mT = \support (\x)
= \{i|x_i \neq 0\} \subset \{1, ..., m\}$ be its support, $\x_{\mT}$
be the subvector indexed by $\mT$, and $\mT^{c}$ be the
complementary set of $\mT$, i.e. $\mT^{c} = \{1, ..., m\}\backslash
\mT$. Furthermore, let $\bA\odot \bB$ represent the element-wise
product of two matrices $\bA$ and $\bB$. Lastly, let $\bA_{\mI}$
denote the columns of $\bA$ indexed by $\mI$.

\subsection{Matching Pursuit LASSO}
To introduce MPL, in \cite{TanPMLPart1}, we bring in a support detection vector
$\dd \in \{0, 1\}^m$ to $\x$, and impose an
$\ell_0$-norm constraint on $\dd$, namely $\|\dd\|_0 \leq \varrho$, to enforce the
sparsity. Here, $\varrho$ is a predefined integer satisfying $1 \leq
\varrho < k$.\footnote{Interested readers may find more
discussions of $\varrho$ in Part I \cite{TanPMLPart1}.} Let $\Lambda = \{\dd: \|\dd\|_0 \leq \varrho, \dd \in
\{0, 1\}^m\}$ be the domain of $\dd$, we propose to solve an integer programming model of LASSO:
\begin{eqnarray}\label{ep:BP_VAR}
&&\min\limits_{\dd\in \Lambda}\min\limits_{\x, \bxi} ~~
\lambda||\x||_1+\frac{1}{2}||\bxi||^2 \\&&\text{s.t.} ~~~~~\bxi =
\db-\bA (\x\odot \dd). \nonumber
\end{eqnarray}
Rather than solving this problem directly,  we bring in dual
variables $\ba \in \R^n$ to the constraint $\bxi = \db-\bA (\x\odot
\dd)$ w.r.t. any fixed $\dd$, and transform (\ref{ep:BP_VAR}) into a minimax problem by introducing the dual form of the inner problem in (\ref{ep:BP_VAR}):
\begin{eqnarray}\label{eq:dual_BP_VAR}
&&\min\limits_{\dd \in \Lambda} \max\limits_{\ba \in \R^n}~~~
-\frac{1}{2}\|\ba\|^2 + \ba^{\top} \db \\&& \mathrm{s.t.}
~~~\|\ba^{\top} \bA \diag(\dd)\|_\infty \leq \lambda.\nonumber
\end{eqnarray}
Let $$f(\ba,\dd) = \frac{1}{2}\|\ba\|^2-\ba^{\top}\db,
~~\ba \in \mA_{\tau}^{\lambda},$$ where $\mA_{\tau}^{\lambda} = \{\ba: \|\ba^{\top}
\bA \diag(\dd)\|_\infty \leq \lambda, \ba\in [-l, l]^{n}\}$ denotes the
 domain of $\ba$ w.r.t. a feasible $\dd$, and $l>0$ is a large number.
By applying a convex relaxation to (\ref{eq:dual_BP_VAR}), MPL seeks to solve the following convex problem:
\begin{eqnarray}\label{eq:convex}
\min\limits_{\ba \in \R^n,\theta \in \R} ~~\theta,
~~~\mathrm{s.t.}~~~f(\ba, \bfeta) \leq \theta ,~~\forall~\dd \in
\Lambda.
\end{eqnarray}

The details of MPL are presented in Algorithm
\ref{Alg:MPL}. Basically, it iteratively adds a set of active atoms
by worst-case analysis in Step 3, and conducts a master problem
optimization in Steps 4-8. Let $\bg = \bA^{\top}\ba^{t-1}$ and
$\mI_t$ be the index set of the detected atoms at the $t$th
iteration, the worst-case analysis is to update $\mI_{t}$ based on
$\bg$. We find the $\varrho$ atoms with the largest
$|g_j|$, and then record their indices into $\mJ_t$. After that, we
update $\mI_t$ by $\mI_{t} = \mI_{t-1} \cup \mJ_t$. The master
problem optimization from Steps 4-8 is to solve the following
problem:
\begin{eqnarray}\label{eq:primal_MPL}
\min_{\x, \bxi}~ \lambda ||\x||_1 + \frac{1}{2}||\db - \bA\x||^2,
~\text{s.t.} ~~\x_{\mI_t^c} = \0.
\end{eqnarray}
The proximal gradient (PG) \cite{Nesterov2007} (resp.
conjugate gradient descent (CGD) \cite{Beckermann2002}) is adopted
to solve (\ref{eq:primal_MPL}) when $\lambda
>0$ (resp. $\lambda = 0$), as shown in the inner \textbf{for loop}.
For the \textbf{for loop}, to distinguish it from the outer \textbf{while loop}, we
use $\bu$ as variables.

\begin{algorithm}[h]
\caption{Matching Pursuit Lasso for Solving (\ref{eq:convex})}
\label{Alg:MPL}
\begin{algorithmic}[1]
\STATE  Initialize $\x^0 = \0$, $\bxi^0 = \db$,  $\mI_0 = \emptyset$.
Let $t = 1$. \WHILE {(The stopping condition is not achieved)}
\STATE Do worst-case analysis: \\Let $\bg = \bA^{\top}\ba^{t-1}$;
choose the $\varrho$ largest $|g_j|$ and record their indices in
$\mJ_t$; let $\mI_{t} = \mI_{t-1} \cup \mJ_t$.\\
\STATE Initialize {$\bu_{\mI_t}^0 = \x_{\mI_{t}}^{t-1}$} and $\bu_{\mI_t^c}^0 = \0$. \\
 \FOR{$s= 1, ...,s_{\max}$}
\STATE Update $\bu_{\mI_t}^s$ using {PG} ($\lambda>0$) or {CGD} ($\lambda=0$) rules. \\
\STATE Break if  the stopping conditions are achieved.\ \ENDFOR.
\STATE Set $\x_{\mI_{t}}^{t} = \bu_{\mI_t}^{k}$, $\x_{\mI_{t}^c}^{t}
= \0$ and $\bxi^t = \db - \bA_{\mI_{t}}\x_{\mI_{t}}^{t}$.   Let
$t = t+1$. \ENDWHILE
\end{algorithmic}
\end{algorithm}

When $\lambda = 0$
and $\varrho = 1$, MPL in Algorithm 1 is reduced to the orthogonal matching pursuit
(OMP)~\cite{pati1993orthogonal,tropp2004greed}. MPL is also related to stagewise
OMP (StOMP)~\cite{donoho2012sparse} and stagewise weak gradient
pursuits (SWCGP for short)~\cite{blumensath2009stagewise}, in the sense that all of them add a set of new atoms per iteration. However, in SWCGP and StOMP, the number of atoms added per iteration changes due to complex thresholding strategies~\cite{donoho2012sparse,blumensath2009stagewise}. For example, in StOMP,
the knowledge of noise is
required to determine the number of new atoms. This knowledge, however, is not available for general
problems~\cite{blumensath2009stagewise}. To address this, SWCGP adopts a
simpler thresholding strategy that is independent of the noise~\cite{blumensath2009stagewise}.
However,  in SWCGP, only one iteration is conducted (namely $s=1$) in the master
problem optimization. As a result, the master problem may not be
sufficiently optimized, and many non-support atoms might be
included accordingly, leading to degraded performance. In contrast, MPL takes more
iterations in the master problem optimization before the following stopping condition is achieved:
\begin{eqnarray}\label{eq:inner}
\frac{f(\bu^{s-1}) - f(\bu^{s})}{f(\bu^{0}) - f(\bu^{s})}
 \leq
\varepsilon_{in},\end{eqnarray} where $\varepsilon_{in}$ denotes a
small tolerance.

\subsection{Subspace Exploratory Matching for MPL}\label{sec:SMPL}

The convergence of MPL has been studied in Part I~\cite{TanPMLPart1}.
However, the performance of MPL might be affected by the value of
$\varrho$. To explain this, we first present a bound regarding
the progress of objective value per outer loop.

\begin{lemma}\label{lemma:improve}
Let $f(\x) = \|\x\|_1 + \frac{1}{2}\|\bxi\|^2$, $\bg =
\bA^{\top}\bxi^{t-1}$ and $\bu^{1}$ be the starting point regarding
the inner loop. Assume $|g_i|>\lambda$ for $\forall i\in \mJ_{t+1}$,
where $\mJ_{t+1}$ is obtained by Step 3 of Algorithm~\ref{Alg:MPL},
with proper line search in PG, we have:
\begin{eqnarray}
f(\x^{t})  - f(\bu^{1}) \geq
 \frac{1}{2L} \sum_{i\in
\mJ_{t+1}}(|g_i|-\lambda)^2, \nonumber
\end{eqnarray}
where $1/L$ is the step size obtained by the line search in PG.
\end{lemma}

According to Lemma \ref{lemma:improve},  choosing $\varrho$
atoms with the largest $|g_i|$ can only guarantee the best improvement in objective values
after one iteration (e.g. $s=1$) of the inner loop. {However, these $\varrho$ atoms cannot
guarantee the best objective value improvement when more inner iterations (e.g. when $s>1$) are used. In other words, the worst-case analysis in Step 3 might be suboptimal when $s>1$. When $\varrho$ is
relatively large in particular, some non-support atoms that are with large values of $|g_i|$ might be mistakenly added into $\mJ_t$. To address this,  we propose to first include more
 than $\varrho$ (e.g. $\omega\varrho$, where $\omega
> 1$) new atoms with the largest $|g_i|$, and then solve the master problem
in (\ref{eq:primal_MPL}) with all of the selected atoms. Finally, we choose $\varrho$ new atoms that decrease the objective value the most as the
most-active atoms. This scheme, which is referred to as  subspace
exploratory matching, is summarized in Algorithm~\ref{Alg:algo_K}. To improve the efficiency, we adopt a warm-start strategy (see Step 3), and use equation (\ref{eq:inner}) as the stopping condition in the master problem optimization.

\begin{algorithm}[H]
\begin{algorithmic}[1]
\STATE Given a dictionary $\bA$, $\mI_{t-1}$,  $\ba^t$, $\varepsilon_{in}$ and $\omega
(\omega \geq 1)$. \STATE Calculate $\bg = \bA^{\top} \ba^t$; choose
the $\omega \varrho$
 largest $|g_{j}|$ and record the indices in
$\mJ_{\omega}$; let $\mI_{\omega} = \mI_{t-1} \cup \mJ_\omega$.
\STATE Initialize {$\bu_{\mI_\omega}^0 = \x_{\mI_\omega}^{t-1}$} and $\bu_{\mI_\omega^c}^0 = \0$. \\
 \FOR{$s= 1, ...,s_{\max}$}
\STATE Update $\bu_{\mI_\omega}^s$ using {PG} ($\lambda>0$) or {CGD} ($\lambda=0$) rules. \\
\STATE Quit if  the stopping conditions  are achieved.\ \ENDFOR.
\STATE Sort the
$\omega \varrho$ atoms in $\mJ_\omega$ in descending order by
$|u_i|$; return  the first $\varrho$ atoms and record the indices in
$\mJ_t$.\\
\STATE Let $\mI_{t} = \mI_{t-1} \cup \mJ_t$. Set $\x_{\mI_{t}}^{t} =
\bu_{\mI_t}^{s}$ and $\x_{\mI_{t}^c}^{t} = \0$.
\end{algorithmic}\caption{\small{Subspace Exploratory Matching}} \label{Alg:algo_K}
\end{algorithm}

For convenience, hereafter we refer to Algorithm \ref{Alg:MPL} with
the subspace search as SMPL. In general, since the $\varrho$ atoms
chosen in SMPL achieve better improvement in objective value than
MPL, both convergence speed and sparse recovery
performance can be boosted, which can be observed in Fig.
\ref{coverge_compare_zero_one} in Section \ref{exp:PGH}.

The proposed subspace search is related to the atom selection strategies used in
CoSaMP~\cite{Needell2009}, SP~\cite{Dai2009} and
{OMPR}~\cite{Jain2011}. \revise{For
example, to find $k$ true supports, CoSaMP and SP choose $2k$ and $k$ additional atoms respectively into the active atom set. After that, a pruning step is performed such that only $k$ atoms are kept in the active atom set. In contrast,  there is no atom replacement or deletion in (S)MPL w.r.t. the outer iterations.  Consequently, SMPL is
guaranteed to monotonically decrease the objective values as in MPL~\cite{TanPMLPart1}. Lastly, the subspace search of CoSaMP, SP and {OMPR}
relies on the estimation of $k$, which  is not
required in SMPL.}
\subsection{Stopping Conditions}
Given a properly selected
$\lambda$, a natural stopping condition for (S)MPL  is
\begin{eqnarray}
\|\ba^{\top}\bA\|_{\infty} \leq \lambda.
\end{eqnarray}
However, in practice, we may choose a small $\lambda$ in order to reduce the solution bias of LASSO directly.
When $\lambda$ is very small, \revise{(S)MPL stops when $||\ba|| \ll ||\be||$ (here $\be$ denotes the
ground-truth noise), and it is possible that the over-fitting problem will happen. To prevent from the over-fitting problem, we
stop (S)MPL early if the following stopping conditions are achieved:}
\begin{eqnarray}\label{eq:stop1}
||\ba^{\top}\bA||_{\infty} \leq r_{\infty} ~~\textrm{or}~~||\ba||
\leq r_{2},
\end{eqnarray}
 \noindent where $r_{\infty}$ and $r_{2}$ are
pre-determined parameters.
We can also stop (S)MPL if
\begin{eqnarray}\label{eq:stop_con}
\frac{\delta^t}{|\varrho f(\x^0)|} \leq \varepsilon,
\end{eqnarray}
where $\delta^t$ is the function value difference between the
$(t-1)^\textrm{th}$ and $t^\textrm{th}$ iteration, $\varepsilon$ is
a small tolerance and $f(\x^0)$ denotes the initial objective
value.

\revise{Without early stopping, (S)MPL will achieve the  LASSO solution, which may be biased (when $\lambda$ is large) or over-fitted (when $\lambda$  is small). For $\lambda=0$ and $\varrho = 1$ in particular, (S)MPL will get the results of OMP~\cite{pati1993orthogonal,tropp2004greed}.}
\subsection{Implementation Concerns}
Several implementation techniques can be adopted to improve the efficiency
of (S)MPL. Note that the master problem optimization in (S)MPL is
w.r.t. a small set of atoms only. Let $\mI$ be the index set of
selected atoms. We only need to calculate small scale matrix-vector
products $\bA_{\mI}\x_{\mI}$ and $\bA_{\mI}^{\top}\bxi$. For
convenience, we refer to them
as the partial matrix-vector product (PMVP). Correspondingly, we refer to $\bA\x$ and
$\bA^{\top}\bxi$ as the full matrix-vector product (FMVP).

Firstly, since $|\mI| \ll m$, computing the PMVP (e.g.
$\bA_{\mI}\x_{\mI}$ and $\bA_{\mI}^{\top}\bxi$) is much cheaper than
FMVP (e.g. $\bA\x$ and $\bA^{\top}\bxi$). To fully exploit this
advantage, we store $\bA$ atom by atom in the main
memory so that we can easily retrieve any atoms indexed by $\mI$ using C++
pointers.

Secondly, when dealing with big dictionaries, the cache-to-memory
efficiency is important. For example, the calculations of PMVPs (e.g. $\bA_{\mI}\x_{\mI}$ and
$\bA_{\mI}^{\top}\bxi$) may not be cache-to-memory efficient, since the
active atoms in general are very far away from each other in the
main memory.  To address this, we explicitly store $\bA_{\mI}$ and
$\bA_{\mI}^{\top}$ in the main memory. Accordingly, we can compute
PMVPs very efficiently.

Thirdly,  several iterations
regarding the master problem optimization  are sufficient, which significantly reduce the number of PMVPs. Moreover, once updating $\mI_t$, we set $\x_{\mI_{t}}^{t} = \bu_{\mI_t}^{s}$ for the purpose of warm-start (see Step 9 in Algorithm \ref{Alg:algo_K}). In this way, we can significantly improve the efficiency of the master problem optimization.\footnote{For fair comparison, we employ the above techniques
to implement the $\ell_1$-norm methods whenever the intermediate
variables are sparse:  Let ${\mI}$ denote the
supports of an intermediate $\x$, we replace $\bA\x$ with
$\bA_{\mI}\x_{\mI}$,
which will improve the efficiency considerably.}

\section{Batch Mode {MPL}}~\label{sec:batch}

In the batch SR problem, suppose there are $p$ signals to be
sparsely represented at the same time. Existing $\ell_1$-norm
methods, such as PG~\cite{Nesterov2007} and FISTA~\cite{Beck2009},
take $O(mn)$ cost per iteration. Suppose they stop after $S$
iterations, the total cost for recovering $p$ signals is $O(Spmn)$.
On the contrary, suppose (S)MPL stops after $T$ iterations, it will reduce
the cost to $O(Tpmn)$, where  $T \ll S$.

Nevertheless, the complexity of MPL and SMPL is still dependent on $n$, making them expensive to tackle large-scale problems that are with large $n$. Essentially, this computational burden is brought
by the calculation of $\bA^{\top}\bxi$ (which takes $O(mn)$ cost) in
the worst-case analysis. Therefore, how to reduce the cost of
$\bA^{\top}\bxi$ is critical for improving the  efficiency.

According to the studies in~\cite{Figueiredo2007,donoho2012sparse},
if the discrete Fourier transform basis or wavelet basis are sampled
to form the dictionary $\bA$, the computational complexity of
$\bA^{\top}\bxi$ can be reduced to $O(m\log(m))$ with the help of
the fast Fourier transform (FFT). However, this technique cannot be
applied to general dictionaries.

{To tackle  many signals under general dictionaries, we propose below the \textbf{batch-mode MPL} (BMPL for
short), in which the computational cost can be greatly reduced.   Actually, we have
$\bA^{\top}\bxi  = \bA^{\top}(\db-\bA_{\mI}\x_{\mI}) = \bA^{\top}\db - [\bA^{\top}\bA_{\mI}] \x_{\mI}.$  Let $\bb = \bA^{\top}\db$ and $\Q =
\bA^{\top}\bA$. If we pre-compute $\Q$ and $\bb$, and store them in
the main memory, we can then calculate $\bA^{\top}\bxi$ according to
\begin{eqnarray}\label{eq:batchQ}
\bA^{\top}\bxi = \bb - \Q_{\mI} \x_{\mI}.
\end{eqnarray}
As a result, the computation cost of computing $\bA^{\top}\bxi$ is
reduced to $O(m|\mI|)$, where $|\mI|\ll n$. Since $|\mI| \approx k$, the overall cost for $p$
signals becomes $O(Tpmk)$.
\begin{remark}
To apply (\ref{eq:batchQ}), we need to
compute the matrix $\Q \in \R^{m\times m}$ with $O(nm^2)$
cost, which is not efficient regarding a single signal. However, since
$\Q$ can be calculated off-line, this cost is negligible when dealing
with many signals.
\end{remark}
Since BMPL adds $\varrho$ atoms per iteration, it requires
considerably fewer times of $\bA^{\top}\bxi$ than the batch mode
OMP (BOMP for short)~\cite{Rubinstein2008}. Specifically, BOMP takes
$O(pmk^2)$ cost for $p$ signals; while BMPL takes
$O(Tpmk)$ complexity, where $T\ll k$.

For existing $\ell_1$-norm methods,  even though the
intermediate variables are sparse, it is not easy for them to
conduct the batch mode optimization, since the support set $\mI$ of intermediate variables might change
frequently during the optimization. As a result, frequent retrievals
of $\Q_{\mI}$ are very computationally expensive. 

The batch scheme is not applicable to a dictionary with a very large number
of atoms, because of the $O(m^2)$ space complexity to store $\Q$. Nevertheless, BMPL can be applied to many large-scale tasks. For example, it can efficiently
deal with dictionaries of $O(2^{15})$ atoms on a 24GB memory machine, which is sufficient for many real-world applications, such as face recognition~\cite{Wright2009} and dictionary learning~\cite{ICML2011Coates}.

\section{Numerical Experiments}\label{sec:num_exp}
In this section, we compare the performance of (S)MPL with
the following baseline methods:\footnote{The C++ source codes of MPL and the compared methods
are available at:
{\url{http://www.tanmingkui.com/mpl.html}.}}
\begin{itemize}
\item
Four state-of-the-art $\ell_1$-solvers:
{Shotgun}\footnote{\url{https://www.select.cs.cmu.edu/projects}.}
which uses the parallel coordinate descent in
C++ \cite{Bradley2011Parallel}.
{FISTA}\footnote{\url{https://www.eecs.berkeley.edu/~yang/software/l1benchmark/index.html}.}
which uses the accelerated proximal gradient method with continuation
technique~\citep{Figueiredo2007,Yang2010ICIP};
{PGH} which uses the homotopy method to improve the convergence
speed~\citep{linxiao,Xiao2013PGH_J};
{S-L1}\footnote{\url{http://www.princeton.edu/~zxiang/home/index.html}.}
which adopts a screening test to predict the zero entries to improve the
decoding efficiency~\cite{James2011}.
\item
Several related greedy methods, such as
ROMP~\cite{Needell2009ROMP}\footnote{\url{https://www-personal.umich.edu/~romanv/software/romp.m}.},
StOMP~\cite{donoho2012sparse}\footnote{\url{https://sparselab.stanford.edu/}.}
and SWCGP~\cite{blumensath2009stagewise} are used for the comparison. In
addition, four well-known greedy algorithms, i.e. orthogonal matching
pursuit {(OMP)}~\cite{pati1993orthogonal,tropp2004greed},
accelerated iterative hard thresholding (AIHT)~\cite{blumensath2009iterative,Blumensath2011,Giryes2011}\footnote{\url{https://www.personal.soton.ac.uk/tb1m08/publications.html}.},
subspace pursuit {(SP)}~\cite{Dai2009}\footnote{\url{https://sites.google.com/site/igorcarron2/cscodes}.}
and orthogonal matching pursuit
with replacement ({OMPR})~\cite{Jain2011},  are also included as baseline methods.
\end{itemize}

\begin{figure*}[htp]
\center
    {\includegraphics[trim = 2mm 6.5cm 2cm 1.4cm,  clip,  width=0.72\textwidth]{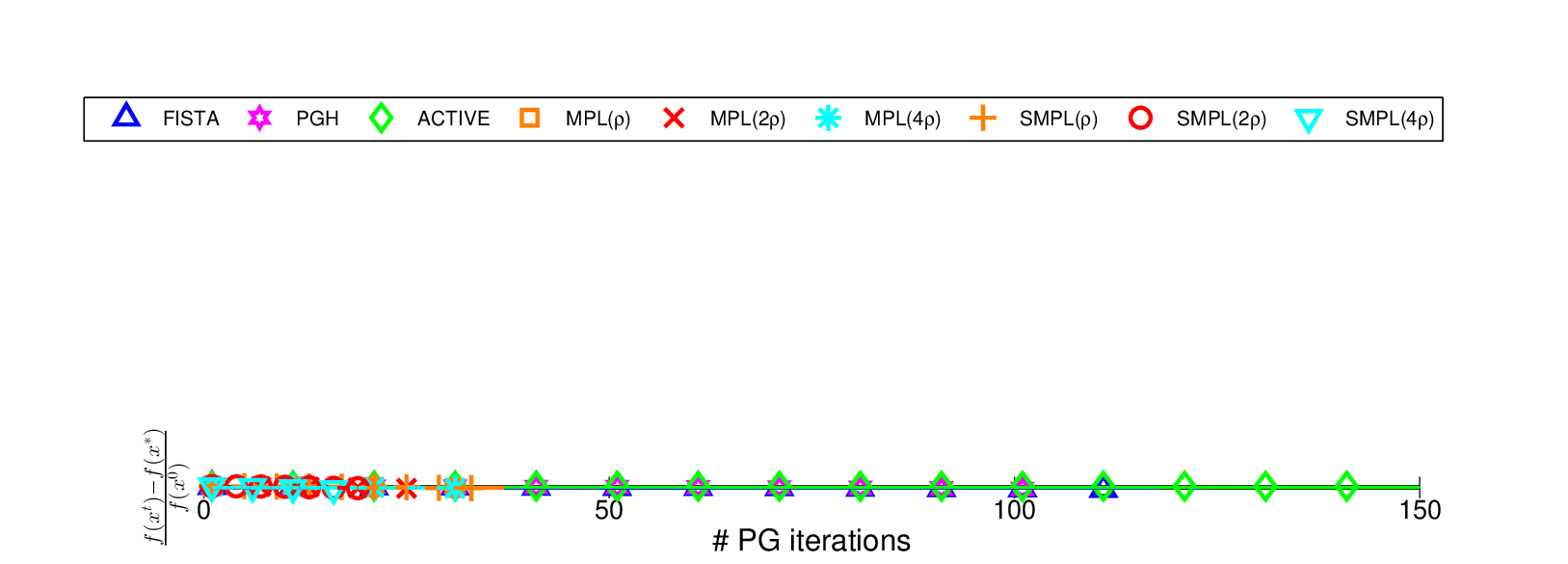}}\vspace{-6mm} \\
    \center
 \subfigure[ {Objective value  evolutions for $\lambda=0.005||\bA^{\top}
\db||_{\infty}$}]{
    \label{zero_one_lambda}
    \includegraphics[trim =2mm 0mm 6mm 0mm, clip,height=2.10in,  width=2.65in]{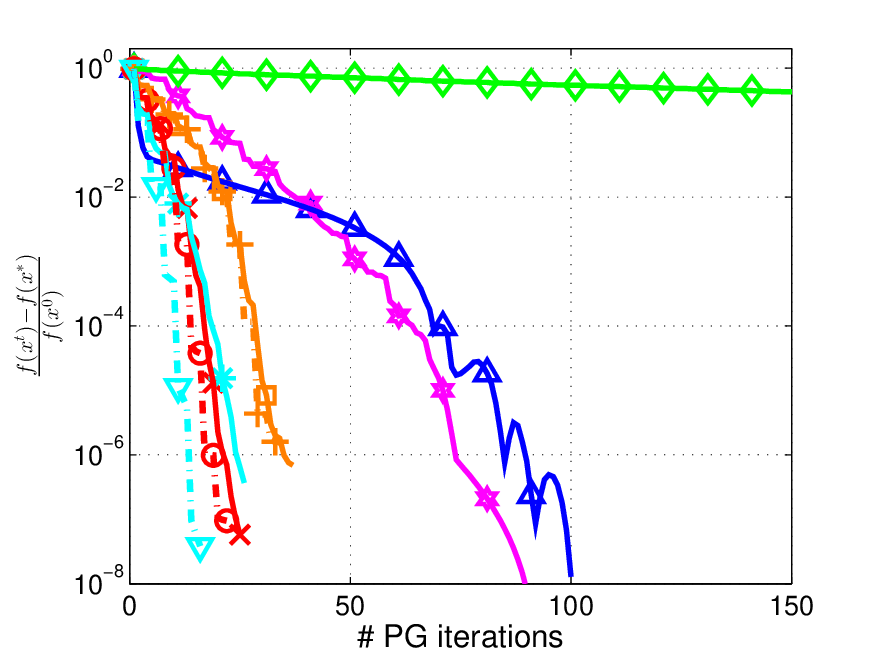}}\hspace{0.2in}
    \subfigure[Objective value  evolutions for $\lambda=0.00005||\bA^{\top}
\db||_{\infty}$]{
    \label{zero_one_lambda_small}
   \includegraphics[trim = 1mm 0mm 6mm 0mm,  clip, height=2.10in,  width=2.65in]{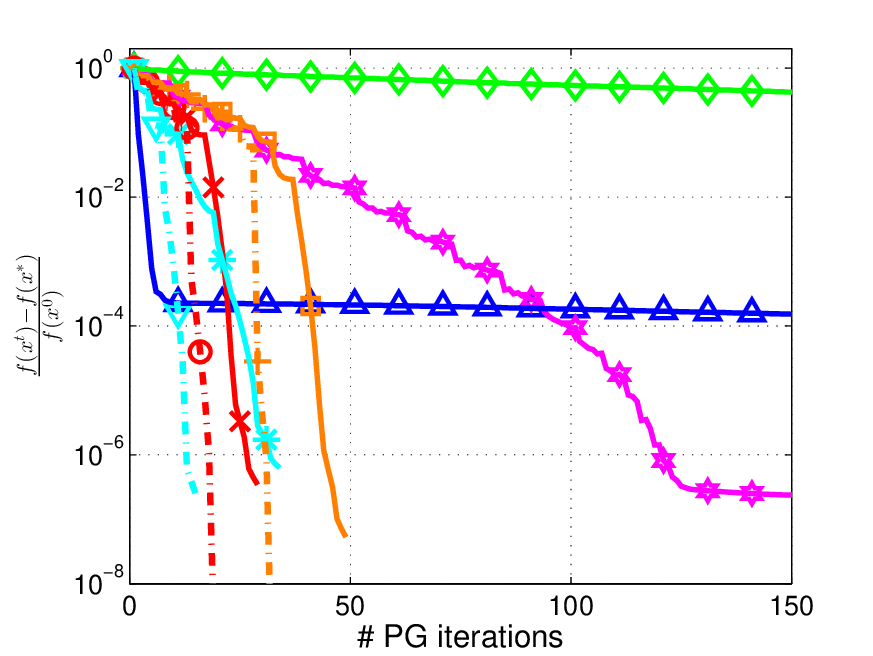}}
 \subfigure[ {Objective value  evolutions for $\lambda=0.005||\bA^{\top}
\db||_{\infty}$}]{
    \label{gauss_lambda}
    \includegraphics[trim =2mm 0mm 6mm 0mm, clip,height=2.10in,  width=2.65in]{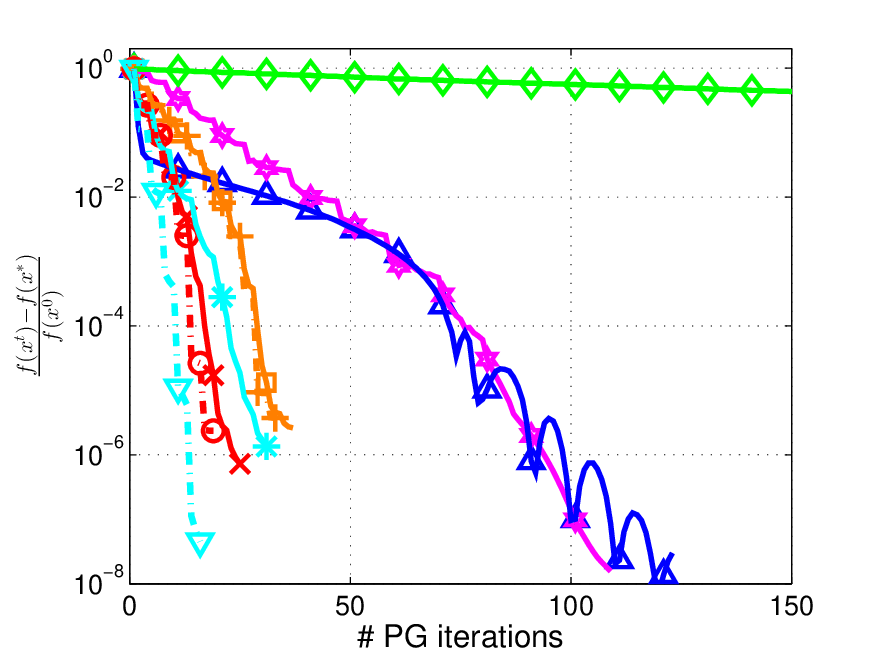}}\hspace{0.2in}
    \subfigure[Objective value  evolutions for $\lambda=0.00005||\bA^{\top}
\db||_{\infty}$]{
    \label{gauss_lambda_small}
   \includegraphics[trim = 1mm 0mm 6mm 0mm,  clip, height=2.10in,  width=2.65in]{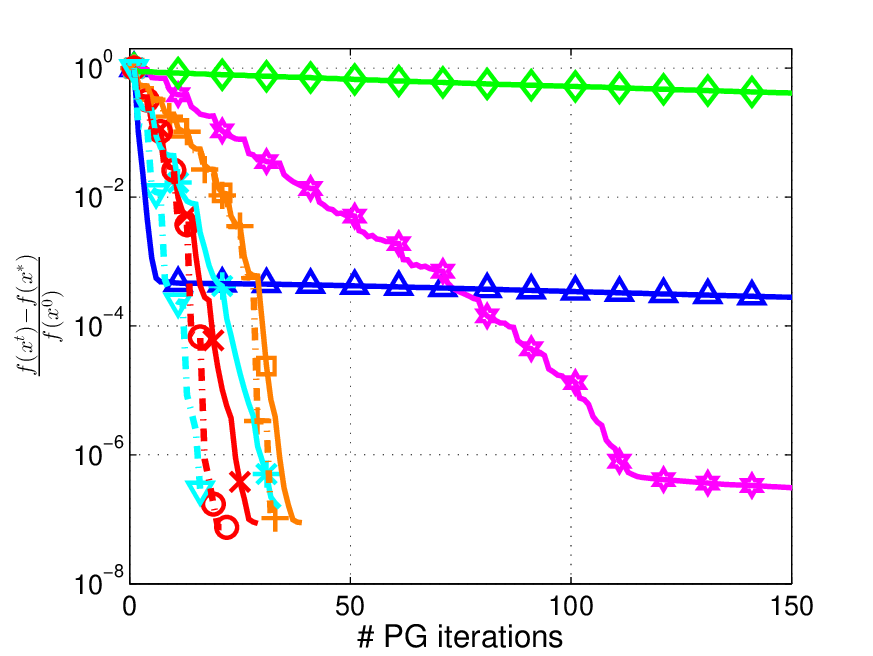}}
  \caption{Convergence  of the comparison methods  on \emph{{Bernoulli}}  sparse
  vectors (in Fig. \ref{zero_one_lambda} and \ref{zero_one_lambda_small}) and \emph{{Gaussian}}  sparse
  vectors (in Fig. \ref{gauss_lambda} and \ref{gauss_lambda_small}). For (S)MPL and the Active-set method, we record $f(\x^t)$ per  PG iteration.  We only record results within 150 iterations for all methods.} \label{coverge_compare_zero_one}
\end{figure*}

In the experiments, {Shotgun} is conducted in parallel on an Intel(R) Core(TM) i7
CPU (8 cores) PC with 64-bit Linux OS; while the other
methods are conducted on a 64-bit Windows operating system (OS) with
the same computer configuration.  For fair comparison, all methods, except
{S-L1}, ROMP and StOMP, {are} written in C++ running with
\textbf{single core}. We run {S-L1}, which is written in Matlab, in parallel on an eight-core machine.

\begin{table*}[htp]
\center \caption{{Comparison among {MPL}, FISTA, PGH and Active-set
methods on \emph{{Bernoulli}} sparse signal, where  \emph{Time} records the
decoding time (in seconds).}}\label{table:acc_PGH}
\begin{scriptsize}
\begin{tabular}{|c|c|c|c|c|c|c|c|c|c|c|}
\hline
       $\lambda$    &           & Active-set &  FISTA &      PGH &  MPL($\rho$) & SMPL($\rho$) &  MPL($2\rho$) & SMPL($2\rho$) &  MPL($4\rho$) & SMPL($4\rho$) \\
\hline \multirow{6}{*}{$0.005||\bA^{\top}
\db||_{\infty}$}      
 &   \emph{Sparsity} &     160 &        595 &        253 &        159 &        168 &        216 &        166 &        188 &        178 \\
\cline{2-11}
\multicolumn{ 1}{|c|}{} & \#FMVP &             160 &        120 &        177 &         11 &         11 &          7 &          5 &          4 &          3 \\
\cline{2-11}
\multicolumn{ 1}{|c|}{} &   \#PMVP &       2591 &        344 &        344 &        228 &        450 &        183 &        238 &        134 &        171 \\
\cline{2-11}
\multicolumn{ 1}{|c|}{} &       \emph{Time}  &    1.36 &      6.07 &      1.22 &       0.11 &       0.14 &      0.09 &      0.08 &       \bf 0.06   &   \bf 0.07 \\
\cline{2-11} \multicolumn{ 1}{|c|}{} &    \#speedup &           21.9 &    97.9 &       19.6 &        1.8 &        2.3 &        1.5 &        1.3 &        1 &        1.3 \\
\hline \multirow{6}{*}{$0.00005||\bA^{\top}
\db||_{\infty}$} 
 &   \emph{Sparsity} &        161 &       1015 &       1015 &        189 &        144 &        244 &        195 &        328 &        197 \\
\cline{2-11}
\multicolumn{ 1}{|c|}{} & \#FMVP &       160 &       1000 &        160 &         13 &         10 &          8 &          6 &          5 &          3 \\
\cline{2-11}
\multicolumn{ 1}{|c|}{} &  \#PMVP  &       2647 &       3021 &        473 &        279 &        418 &        202 &        281 &        175 &        170 \\
\cline{2-11}
\multicolumn{ 1}{|c|}{} &       \emph{Time} &    1.47 &     98.37 &      2.78 &      0.12 &      0.14 &      0.09 &      0.09 &   \bf   0.08 &     \bf 0.08 \\
\cline{2-11}
\multicolumn{ 1}{|c|}{} &   \#speedup &     18.8 &   1261.2 &       35.6 &        1.6 &        1.8 &        1.2 &        1.2 &        1.0 &        1.0 \\
\hline
\end{tabular}
\end{scriptsize}
\end{table*}

\begin{table*}[htp]
\center \caption{{Comparison among {MPL}, FISTA, PGH and Active-set
methods on \emph{Gaussian} sparse signal, where  \emph{Time} records the
decoding time (in seconds).}}\label{table:acc_PGH_gauss}
\begin{scriptsize}
\begin{tabular}{|c|c|c|c|c|c|c|c|c|c|c|}
\hline
       $\lambda$    &&     Active-set & FISTA &  PGH &  MPL($\rho$) & SMPL($\rho$) &  MPL($2\rho$) & SMPL($2\rho$) &  MPL($4\rho$) & SMPL($4\rho$) \\
\hline \multirow{6}{*}{$0.005||\bA^{\top} \db||_{\infty}$}& 
  \emph{Sparsity} &        160 &        313 &        221 &        154 &        154 &        196 &        168 &        303 &        256 \\
\cline{2-11} \multicolumn{ 1}{|c|}{} & \#FMVP &            160 & 79
&         92 &         10 &         10 &          6 &          5 & 5
&          4 \\
\cline{2-11}
\multicolumn{ 1}{|c|}{} &   \#PMVP &       2578 &        255 &        280 &        195 &        414 &        146 &        235 &        152 &        208 \\
\cline{2-11}
\multicolumn{ 1}{|c|}{} &       \emph{Time} &        1.40 &      4.46 &      0.92 &      0.09 &       0.14 &     \bf 0.06 &      0.09 &     \bf 0.06 &      0.09 \\

\cline{2-11}
\multicolumn{ 1}{|c|}{} &   \#speedup &    22.3 &   70.8 &       14.6 &        1.5 &        2.2 &        1.0 &        1.5 &        1.0 &        1.5 \\
\hline \multirow{6}{*}{$0.00005||\bA^{\top}
\db||_{\infty}$} 
 &   \emph{Sparsity} &      160 &       1015 &        1015 &        166 &        154 &        222 &        194 &        391 &        280 \\
\cline{2-11}
\multicolumn{ 1}{|c|}{} & \#FMVP  &      201 &       1000 &        144 &         11 &         10 &          7 &          6 &          6 &          4 \\
\cline{2-11}
\multicolumn{ 1}{|c|}{} &  \#PMVP &       3271 &       3023 &        611 &        238 &        415 &        183 &        282 &        202 &        239 \\
\cline{2-11}
\multicolumn{ 1}{|c|}{} &       \emph{Time} &         1.92 &     92.41 &      2.28 &      0.12 &      0.12 &     \bf 0.09 &     \bf 0.09 &   \bf   0.09 &      0.11 \\
\cline{2-11}
\multicolumn{ 1}{|c|}{} &   \#speedup &          20.5 &    983.1 &       24.2 &        1.3 &        1.3 &        1.0 &        1.0 &        1.0 &        1.2 \\
\hline
\end{tabular}
\end{scriptsize}
\end{table*}

\subsection{Experimental Settings and Performance Metrics}\label{sec:param_setting}
Following~\cite{linxiao,Figueiredo2007}, we set $\lambda = 0.005||\bA^{\top} \db||_{\infty}$ for $\ell_1$-norm methods. Unless noted otherwise, we apply  \textbf{de-biasing technique} to
reduce the solution bias of $\ell_1$-norm methods~\citep{Figueiredo2007,Yang2010ICIP}.
For (S)MPL,
we apply the early stopping to avoid the over-fitting problem with
stopping condition
\begin{eqnarray}\label{eq:stop}
{{|\delta^t|}/({\varrho||\db||^2}}) \leq 1.0 \times 10^{-5},
\end{eqnarray}
where $\delta^t$ denotes the objective difference between the $t$th
and $(t+1)$th iterations. We set the subspace
search length $\omega = 3$ for SMPL.  For many greedy methods, such as AIHT, SP and OMPR, we need to specify $\widehat{k}$. In the simulation, since we know the ground-truth $k$, we set $\widehat{k} = 1.2{k}$. For OMPR, $\eta$ is
set to 0.7. Lastly, we keep default settings of other
parameters for the baseline methods.

Following \citep{Dai2009,linxiao,Xiao2013PGH_J}, we study compressive sensing
problems over \emph{Gaussian} design matrices. We study two types of
sparse signals, e.g. \emph{{Bernoulli}} sparse vector (denoted by
$\s_z$ with each nonzero entry being either 1 or -1) and
{\emph{Gaussian}} sparse signal (denoted by $\s_g$ with each nonzero entry
being sampled from \emph{Gaussian} distribution $\mN(0, 1)$). The observation $\db$ is produced by $\db = \bA\x + \be$,
where $\be$ denotes the additive noise uniformly sampled from $[-0.01,
0.01]$.

 To evaluate the sparse recovery performance of a method, we
adopt the \emph{root-mean-square error} (RMSE) as the comparison
metric,
$$\textrm{RMSE} = \sqrt{\sum_{i=1}^m{(x_i^*-x_i)^2}/m},$$
where $\x^*$ denotes the recovered signal. Here, a sparse signal is successfully recovered if $\textrm{RMSE}
\leq$1E$-3$. For a complete comparison, we record the \emph{{empirical probability of successful reconstruction}}~(EPSR) over $M$ independent experiments~\cite{Dai2009}.

\subsection{Comparison with PGH, FISTA and Active-set Method}\label{exp:PGH}

We compare (S)MPL with PGH, FISTA and Active-set methods
on recovering a 140-sparse \emph{{Bernoulli}} sparse signal and a 140-sparse
\emph{Gaussian} sparse signal over a \emph{Gaussian} dictionary $\bA \in
\RR^{2^{10} \times 2^{13}}$.  To study the effect of $\varrho$, given a basic $\varrho$, we study
$2\varrho$ and ${4\varrho}$. We study  two
$\lambda$'s, namely $\lambda_1 = 0.005||\bA^{\top} \db||_{\infty}$
and $\lambda_2 = 0.00005||\bA^{\top} \db||_{\infty}$. In Fig. \ref{coverge_compare_zero_one}, we report the objective
values of the comparison methods w.r.t. iterations.
In Table \ref{table:acc_PGH} and Table
\ref{table:acc_PGH_gauss}, we record the following metrics: The number of
full matrix-vector products (\#FMVPs); The number of partial
matrix-vector products (\#PMVPs);  The number of nonzeros
(\emph{Sparsity}) in solutions; The decoding time (\emph{Time})
for each signal; The speedup (\emph{\#speedup}) of the fastest
method over others.

Based on the results, we draw the following
conclusions.


\begin{itemize}
\item
From Fig. \ref{coverge_compare_zero_one},  (S)MPL  with different
$\varrho$'s converge much faster than baseline methods. In particular,
\textbf{SMPL(2$\varrho$) is about \textbf{20} times
faster} than  others on the \emph{Gaussian} sparse
signal. FISTA converges well when $\lambda = 0.005||\bA^{\top}
\db||_{\infty}$. In particular, the objective value decreases very quickly at the beginning. However,
it converges very slowly when $\lambda =
0.00005||\bA^{\top} \db||_{\infty}$. In fact, generally speaking, the convergence rate of FISTA is only sub-linear, e.g.
$O(1/k^2)$~\cite{Beck2009}. In contrast to FISTA, PGH solves a sequence of subproblems, and attain linear convergence rate if the subproblem is strongly convex~\cite{linxiao,Xiao2013PGH_J}. Overall, it performs much better than FISTA.
\item
 Note that each FMVP takes $O(mn)$ complexity. From Tables \ref{table:acc_PGH} and
\ref{table:acc_PGH_gauss}, (S)MPL  with different $\varrho$'s need
far fewer FMVPs than other methods, which explains the significant speedup of (S)MPL over other methods. Therefore, (S)MPL are more suitable for big dictionaries.
\item
From Tables \ref{table:acc_PGH} and \ref{table:acc_PGH_gauss}, in general, (S)MPL also need much fewer number of PMVPs than others. Moreover, the scale of PMVPs in (S)MPL is much smaller than in PGH and FISTA. For example, when $\lambda = 0.00005||\bA^{\top} \db||_{\infty}$, the
sparsity of the PGH solution is 1015, which
is much larger than that of (S)MPL. In other words, the master problem optimization in PGH is more expensive.
\item
If $\varrho$ is too large, MPL may take more
computation time. For example, from Table \ref{table:acc_PGH}, MPL with $2\varrho$ indeed needs less
time than MPL with $4\varrho$. The reason is that, if $\varrho$ is large,  some
non-support atoms might be mistakenly included. From Fig.
\ref{coverge_compare_zero_one}, {SMPL} in general converges faster
than {MPL} with a large $\varrho$, which demonstrates the
effectiveness of the \emph{subspace exploratory search}.

\item
From Tables \ref{table:acc_PGH} and
\ref{table:acc_PGH_gauss}, the recovered signals are not exactly
$140$-sparse. This is because the observation $\db$ has been disturbed by the noises $\be$.
\end{itemize}

\begin{table}[htp]
\center \caption{Averaged sparsity of solutions obtained by various
methods with $k = 140, 160, 180$,
respectively.}\label{table:acc_StOMP}
\begin{scriptsize}
\begin{tabular}{|c|c|c|c|c|c|}
\hline
      $k$ &       ROMP &      StOMP &      SWCGP &        MPL &       SMPL \\
\hline
       140 &        506 &        260 &        230 &        167 &        154 \\
\hline
       160 &        584 &        309 &        359 &        182 &        168 \\
\hline
       180 &        651 &        374 &        432 &        196 &        210 \\
\hline
\end{tabular}
\end{scriptsize}
\end{table}

\begin{figure*}[htp]\vspace{-5pt}
\center
     \subfigure[{EPSR} w.r.t $k$]{
     \label{fig:SMPL_EPSR}
    \includegraphics[trim = 2mm 1mm 6mm 4mm,  clip,height=2.10in,
    width=2.65in]{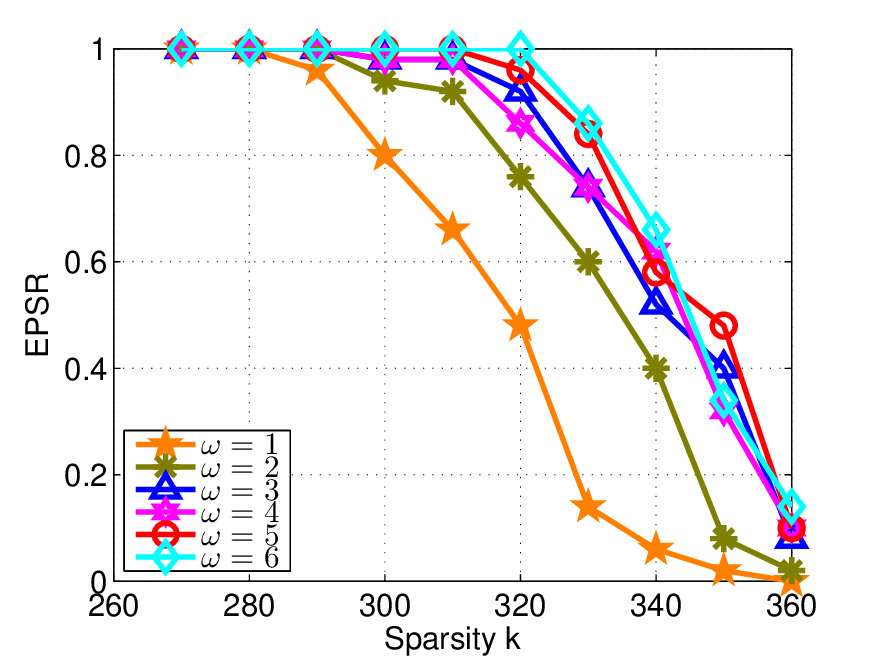}}\hspace{0.2in}
    \subfigure[{Decoding time  w.r.t $k$}]{
    \label{fig:SMPL_time}
    \includegraphics[trim = 2mm 1mm 6mm 4mm,  clip,height=2.10in,
    width=2.65in]{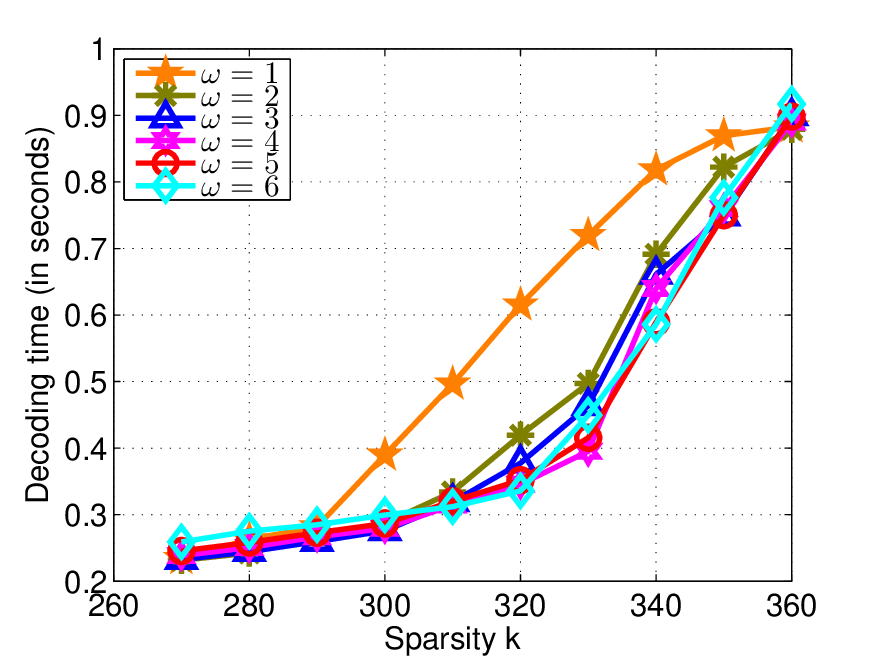}}
  \caption{Results of SMPL on Gaussian sparse signals with different $\omega$'s.}\label{fig:SMPL_omega}
\end{figure*}

\begin{figure*}[htp]\vspace{-5pt}
\center
     \subfigure[{EPSR} w.r.t $k$]{
     \label{fig:StOMP_EPSR}
    \includegraphics[trim = 2mm 1mm 6mm 4mm,  clip,height=2.10in,
    width=2.65in]{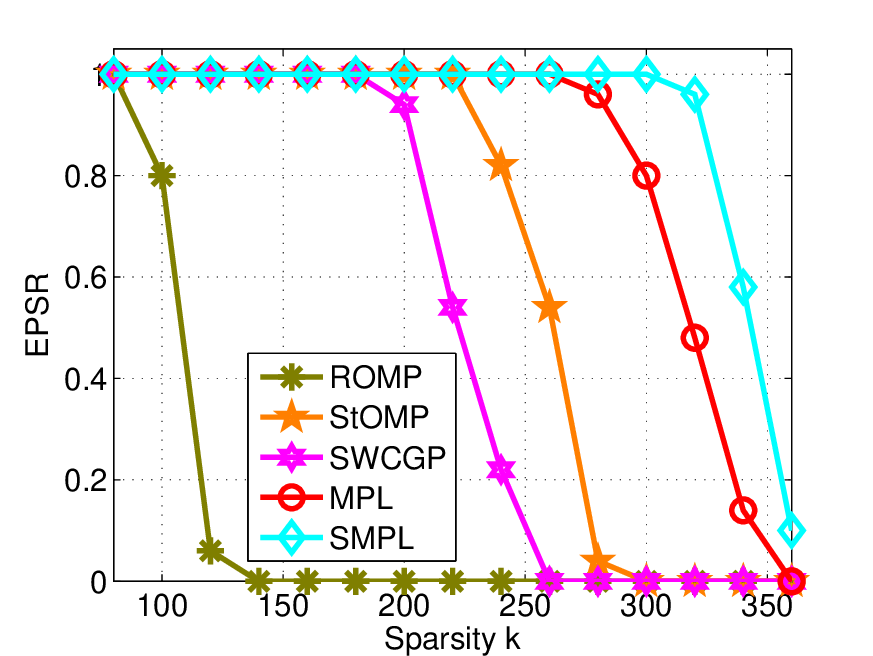}}\hspace{0.2in}
    \subfigure[{Decoding time  w.r.t $k$ (in log scale)}]{
    \label{fig:StOMP_time}
    \includegraphics[trim = 2mm 1mm 6mm 4mm,  clip,height=2.10in,
    width=2.65in]{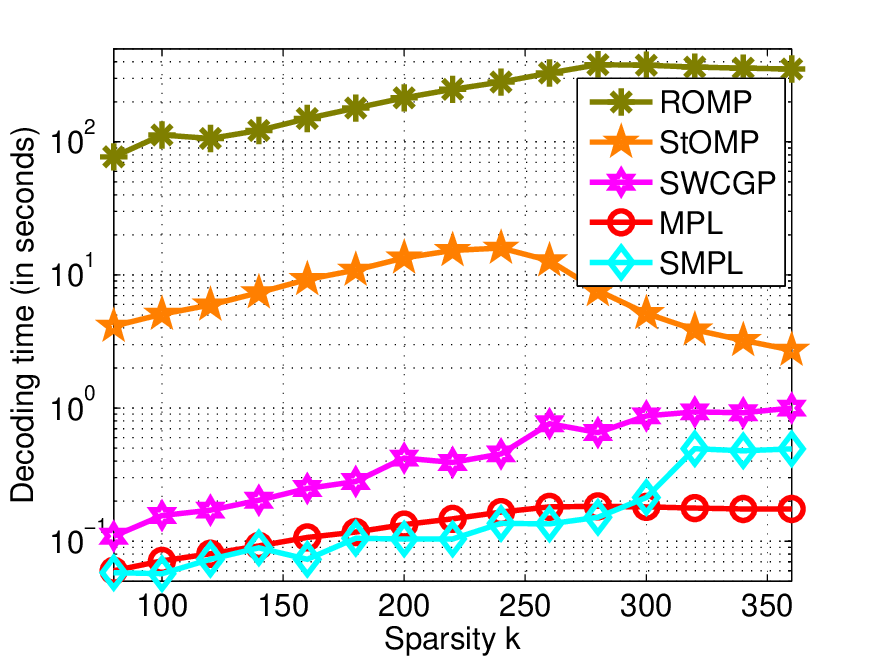}}
  \caption{Comparison among  ROMP, StOMP, SWCGP, MPL and SMPL on Gaussian sparse signals, where the early stopping according to the condition (\ref{eq:stop}) is applied to StOMP, SWCGP, MPL and SMPL. }\label{fig:StOMP}
\end{figure*}

\subsection{Influences of $\omega$ on SMPL}\label{sec:SMPL_omega}
In this experiment, we conduct a sensitivity study on $\omega$ for SMPL. We fix  $\lambda = 0.00005\|\bA^{\top}\db\|_\infty$ and vary $\omega
\in \{1, 2, 3, 4, 5\}$. Note that SMPL is
reduced to MPL when $\omega=1$. For each $k\in \{270, 280, ..., 360\}$, we conduct
$M=100$ independent experiments, and record the EPSR values and averaged
decoding time in Fig. \ref{fig:SMPL_EPSR} and Fig.
\ref{fig:SMPL_time},  respectively.

From Fig. \ref{fig:SMPL_EPSR}, SMPL with larger $\omega$'s
tends to have better recovery performance in terms of EPSR. However,
when $\omega>3$, the improvement becomes less significant. The
reason is that, if $\omega$ is large enough (e.g. $\omega=3$), the
$\omega \varrho$ atoms with largest $|g_i|$ already include most of the
potential active atoms, thus the increasing
$\omega$ will not significantly improve the performance.
From Fig. \ref{fig:SMPL_time},  MPL  (e.g. SMPL
with $\omega = 1$) shows the worst decoding efficiency. The reason is that, without the subspace
search, some non-support atoms might be mistakenly included, and MPL needs more iterations to converge.

\begin{figure*}[htp]
\center
     \subfigure[{EPSR on \emph{Gaussian} sparse signals}]{
    \includegraphics[trim = 2mm 1mm 6mm 4mm,  clip,height=2.10in,  width=2.65in]{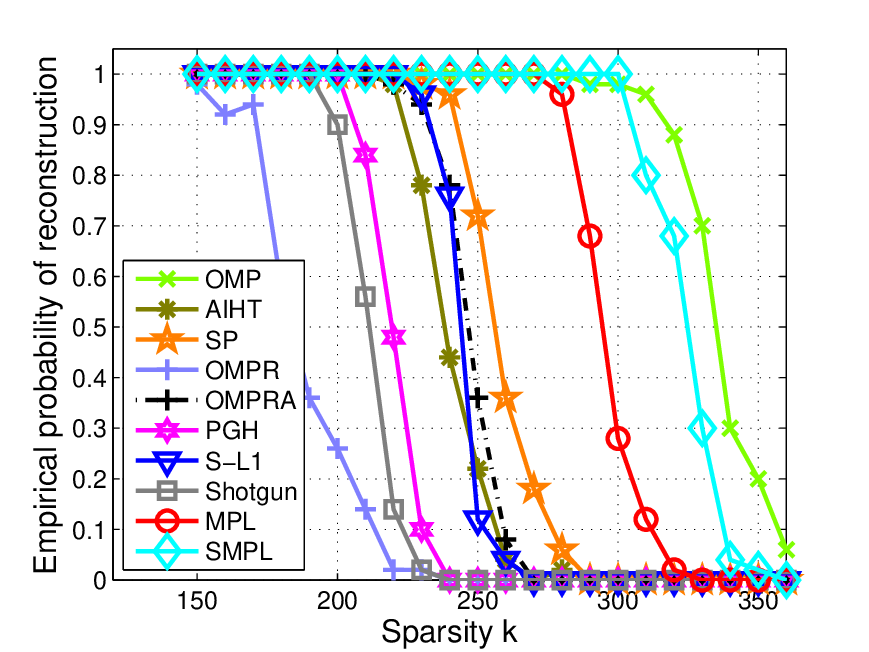}}\hspace{0.2in}
   \subfigure[{Recovery time on  \emph{Gaussian} sparse signals}]{
    \includegraphics[trim = 2mm 1mm 6mm 4mm,  clip,height=2.10in,
    width=2.65in]{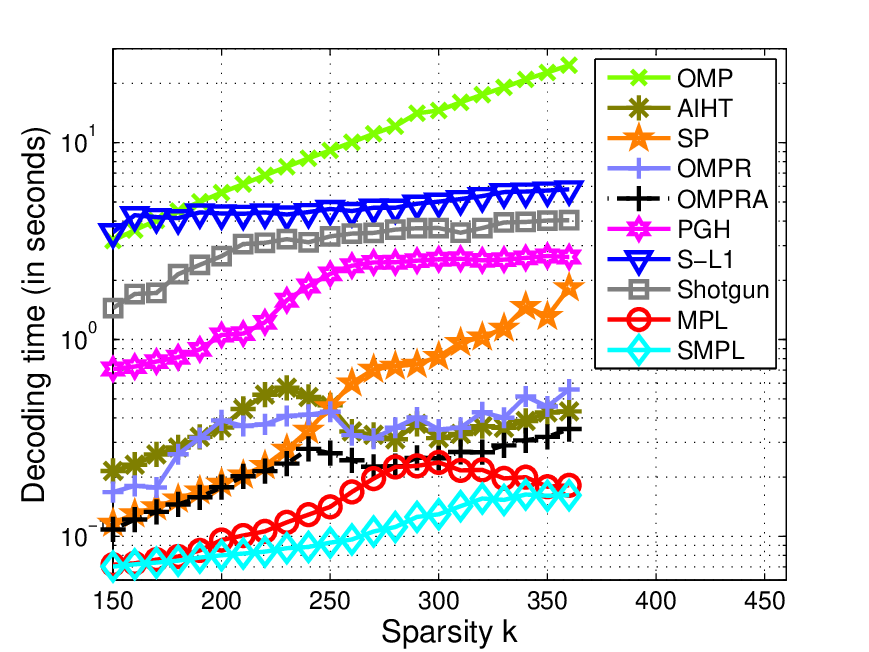}\label{fig:small_time}}
  \caption{SR results on $\bA \in \RR^{2^{10}\times2^{13}}$ of different methods. Here, the de-biasing technique is applied to $\ell_1$-norm methods, and the early stopping is applied to (S)MPL.}\label{fig:small}
\end{figure*}

\subsection{Comparisons with ROMP, StOMP, and SWCGP}\label{sec:SWCGD}
We compare (S)MPL with ROMP,
StOMP, and SWCGP on \emph{Gaussian} sparse signals, where $\bA \in \R^{2^{10}\times2^{13}}$. We use the default parameter settings for
StOMP and SWCGP.
We conduct
$M=100$ independent experiments for each $k \in \{80, 100, ..., 360\}$, and record the EPSR value and the averaged decoding time in Fig. \ref{fig:StOMP_EPSR} and Fig.
\ref{fig:StOMP_time}, respectively. We also record the sparsity
of solutions for $k \in \{140, 160, 180\}$ in Table
\ref{table:acc_StOMP}.

From Fig. \ref{fig:StOMP_EPSR} and Fig.
\ref{fig:StOMP_time}, (S)MPL outperforms the two baselines in terms of sparse recovery
performance and decoding efficiency. StOMP cannot successfully recover all the sparse signals when $k
>240$.  From Table \ref{table:acc_StOMP}, StOMP and SWCGP include more atoms than (S)MPL, which indicates that many
non-support atoms have been included. This problem becomes more
severe for SWCGP, since its master problem is not sufficiently
optimized. As a result, it
cannot recover all the $k$-sparse signals when $k>180$, as shown in Fig. \ref{fig:StOMP_EPSR}. Lastly, ROMP shows much worse
sparse recovery performance than other methods, which is consistent
with the conclusions in~\cite{blumensath2009stagewise}.

\begin{figure*}[htp]\vspace{-5pt}
\center
     \subfigure[{RMSE} w.r.t $k$]{
     \label{fig:large_RMSE}
    \includegraphics[trim = 2mm 1mm 6mm 4mm,  clip,height=2.10in,
    width=2.65in]{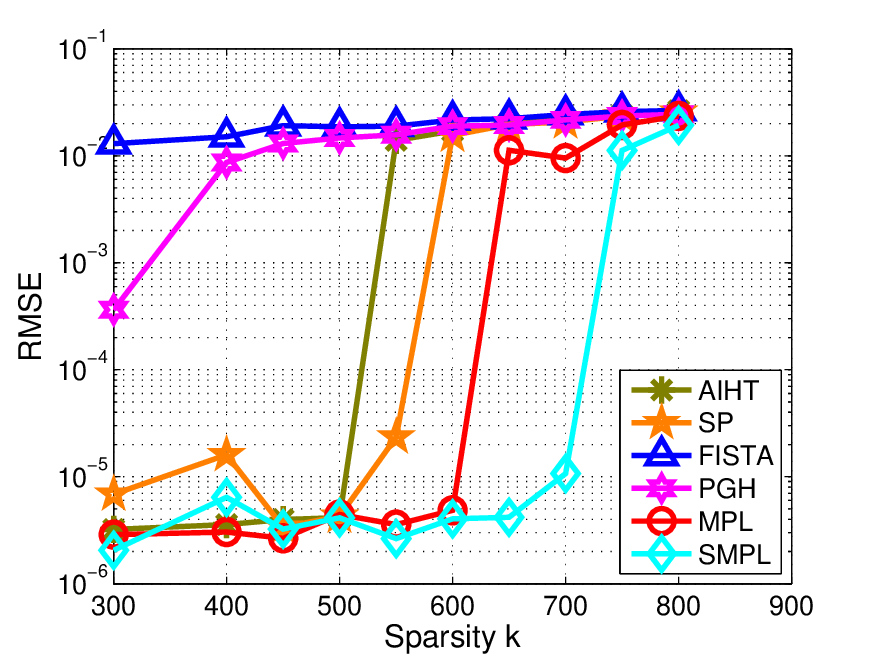}}\hspace{0.2in}
    \subfigure[{Decoding time  w.r.t $k$ (in log scale)}]{
    \label{fig:large_time}
    \includegraphics[trim = 2mm 1mm 6mm 4mm,  clip,height=2.10in,
    width=2.65in]{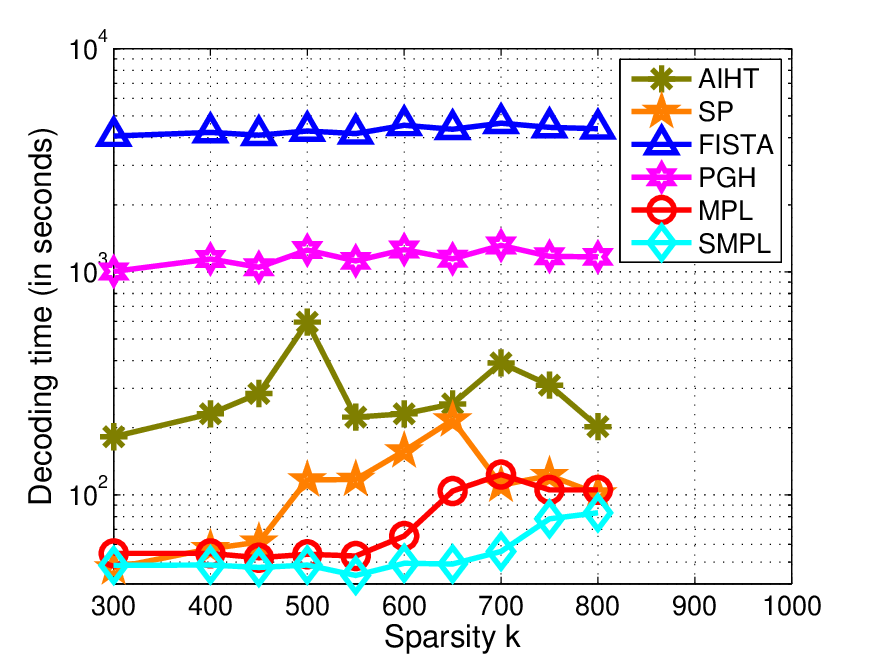}}
  \caption{SR results on Gaussian sparse signals under a \emph{Big Dictionary}  $\bA \in \RR^{2^{12}\times 2^{20}}$.}\label{fig:large_scale}
\end{figure*}

\begin{figure*}[htp]\vspace{-5pt}
\center
     \subfigure[{EPSR} w.r.t $k$]{
     \label{fig:large_EPSR_freq}
    \includegraphics[trim = 2mm 1mm 6mm 4mm,  clip,height=2.10in,
    width=2.65in]{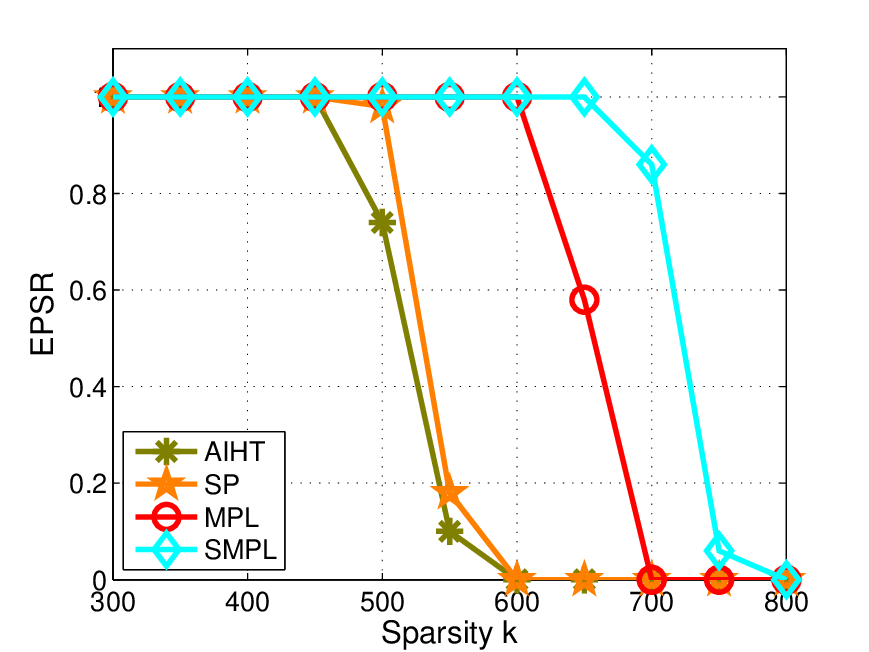}}\hspace{0.2in}
    \subfigure[{Decoding time  w.r.t $k$ (in log scale)}]{
    \label{fig:large_time_freq}
    \includegraphics[trim = 2mm 1mm 6mm 4mm,  clip,height=2.10in,
    width=2.65in]{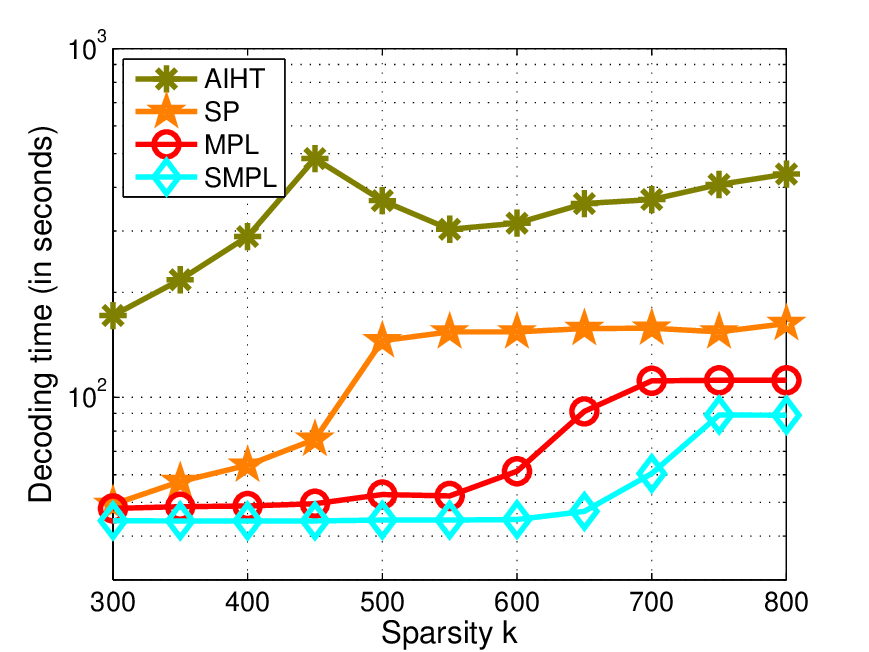}}
  \caption{SR performance comparison under a \emph{Big Dictionary}  $\bA \in \RR^{2^{12}\times 2^{20}}$.}\label{fig:large_scale_freq}
\end{figure*}

\subsection{Comparisons with Other Baselines}\label{sec:expt}
In this experiment, we compare the performance of (S)MPL with other
baseline methods on a median-scale problem $\bA\in
\RR^{2^{10}\times2^{13}}$, where {Shotgun} and {S-L1} work in
{\bf parallel}. For each $k$, we run $M=100$ independent
trials. For (S)MPL, we apply early stopping to avoid the over-fitting problem.

In {OMPR}, it is necessary to calculate $\z = \x + \eta
\bA^{\top}(\db-\bA\x)$, where $\eta$ is a learning
rate of OMPR~\cite{Jain2011}. The setting of $\eta$ is crucial for the
performance~\cite{Jain2011}. In~\cite{Jain2011}, a feasible range
for $\eta$ is provided if $\bA$ satisfies the RIP condition.
Unfortunately, if $\bA$ is not well scaled, the scale of
$\bA^{\top}(\db-\bA\x)$ may vary a lot and the setting of $\eta$
will be difficult.\footnote{{Interested readers can find
more details of $\eta$ in~\cite{Jain2011}.}} To address this issue, we
propose a variant of OMPR in which $\eta$ is adaptively adjusted by
applying the {CGD} rule. \emph{To distinguish this variant from OMPR, we
refer it to as the {OMPRA}}.

The EPSR value and recovery time for the \emph{Gaussian} sparse
signals of each method are presented in Fig.~\ref{fig:small}. From
this figure,  {SMPL} and {OMP} show
much better recovery performance  than other methods on the Gaussian sparse signals in terms of EPSR. In general, {SMPL} shows better recovery performance
than {MPL} in terms of EPSR. {OMPR}~\cite{Jain2011} shows  worse recovery
performance than other greedy methods. From the experiments, {OMPRA}
that uses an adaptive learning rate improves {OMPR} greatly. However, OMPRA
is still worse than {(S)MPL}.

From Fig. \ref{fig:small_time}, {MP} algorithms are much
faster than the $\ell_1$-norm methods, such as {Shotgun} (a
well-designed parallel $\ell_1$-method) and PGH. Ultimately, {PGH}
shows better efficiency than {Shotgun} and {S-L1}, but is much worse than
(S)MPL.

\subsection{Scalability Comparisons on Big Dictionaries}\label{sec:scalability}
{In the final experiment, we compare the scalability of
(S)MPL with several baselines on a Big Dictionary $\bA \in
\RR^{2^{12}\times 2^{20}}$ with two experiments.\footnote{In real-world applications, such as the face recognition task, we may have more than 1 million training images from many persons~\cite{Taigman2011billionface}. In SR based face recognition, the training images are formed as
a big dictionary.} Here, only Gaussian sparse signals are
studied.

In the first experiment, we generate
$k$-sparse signals with $k \in \{300, 400, ..., 800\}$, and compare {(S)MPL} with {FISTA}, {PGH}, {SP} and {AIHT}.  We set $\widehat{k} = 1.2k$ for SP and AIHT. We set $\lambda = 0.005\|\bA^{\top}\db\|_\infty$ for LASSO related algorithms, and set the maximum iterations of FISTA
and PGH to 150.  We report the RMSE and recovery time in Fig. \ref{fig:large_RMSE}
and Fig. \ref{fig:large_time}, respectively. According to the
reported results, the following conclusions can be drawn.}
{\begin{itemize}
\item
From Fig. \ref{fig:large_RMSE}, (S)MPL shows better RMSE than other methods when $500< k\leq 600$; SMPL significantly improves MPL in terms of RMSE when $650
< k < 700$. In addition, SP and AIHT cannot recover the $k$-sparse
signal if $k>600$ (the RMSE values are very large). Lastly, PGH and FISTA
show worse recovery performance than other methods in terms of RMSE, which coincides with the results
in Tables \ref{table:acc_PGH} and \ref{table:acc_PGH_gauss}.
\item
From Fig. \ref{fig:large_time}, it is evident that {(S)MPL} is much more
efficient than other methods, in particular when $k\geq500$.
SP has comparable efficiency with (S)MPL when $k\leq 450$,
but becomes less efficient when $k > 450$. PGH and FISTA need thousands of seconds for all $k$'s; while MPL needs
less than 100 seconds when $k
\leq 600$. In particular, SMPL needs less than 50 seconds  when $k \leq 700$.

\item
From Fig.
\ref{fig:large_RMSE}, it is clear that PGH is better than
FISTA in terms of RMSE. In general, PGH
converges faster than FISTA, thus it achieves a better solution with the same number of iterations.
\end{itemize}}

There are two reasons  for the inefficiency of PGH and
FISTA. {Firstly}, both of them require many iterations to
converge, which means that they need to compute many times of
$\bA^{\top}\bxi$ than (S)MPL.
{Secondly},  when
computing $\bA^{\top}\bxi$ for large dictionaries, the data exchange
between the main memory and cache memory are very inefficient.  In
contrast, in (S)MPL, the master problem optimization is
w.r.t.  a small set of
active atoms only, e.g. $\bA_{\mI}$. Apparently, the data exchange between  the main memory
and cache memory w.r.t. $\bA_{\mI}$ is much more efficient.

To thoroughly compare the scalability of (S)MPL with SP and AIHT, in the second experiment, we run $M=100$ independent
experiments for each $k$, where we exclude FISTA and PGH from the comparison. Here, we set $\widehat{k} = 1.5k$ for SP and AIHT. We record the EPSR value and averaged recovery time in Fig.~\ref{fig:large_EPSR_freq} and Fig.~\ref{fig:large_time_freq}, respectively.
From Fig.~\ref{fig:large_EPSR_freq}, (S)MPL shows much better recovery performance than SP and AIHT in terms of EPSR value. From Fig.~\ref{fig:large_time_freq}, (S)MPL is also much more efficient than SP and AIHT.

\section{Batch MPL and Applications to Many-Face Recognition}\label{exp:batch}
In this section, we first compare BMPL with BOMP
on synthetic compressive sensing tasks, and then apply
them to many-face recognition tasks.

\subsection{Comparison of BMPL and BOMP}\label{exp:batch_simu}

BOMP is a batch mode implementation of
OMP~\cite{Rubinstein2008}. In the simulation, we generate a
\emph{Gaussian} random matrix $\bA \in \R^{2^{12}\times 2^{14}}$ and
generate 200 \emph{Gaussian} sparse signals for each sparsity $k
\in$ from \{400, 450, 500, 550, 600\}. The vector of measurements
$\db$ is produced by $\db = \bA\x + \bxi$ with \emph{Gaussian} noise
sampled from $\mN(0, 0.05)$. The total time (in seconds) spent by
{BMPL} and {BOMP}  in decoding 200 signals and the
\emph{averaged root-mean-square error} (\textbf{ARMSE}) are reported in Table \ref{table:acc_syn2}. From Table
\ref{table:acc_syn2}, {BMPL} is \textbf{about 7-16
times faster} than {BOMP}.  Moreover, BMPL gains better or comparable
\textbf{ARMSE} to {BOMP} for all $k$.

\begin{table}[htp]
\center \caption{Efficiency  {Comparison Between {BMPL} and {BOMP}
(in seconds). The time consumed for computing $\bA^{\top}\bA$ is
46.27 seconds}}\label{table:acc_syn2}
\begin{scriptsize}
\begin{tabular}{|c|c|c|c|c|c|c|}
\hline
           &         $k$ &        400 &        450 &        500 &        550 &        600 \\
\hline
\multirow{2}{*}{BOMP} &       Time &     434.27 &     546.70 &     680.96 &     835.72 &    1014.93 \\
\cline{2-7}
 &        ARMSE &   7.11E-03 &    7.69E-03 &   7.92E-03 &   8.59E-03 &   8.94E-03 \\
\hline
\multirow{3}{*}{BMPL}&       Time &      55.06 &      55.79 &      56.79 &      59.51 &      59.91 \\
 \cline{2-7} &        ARMSE &   3.88E-03 &    4.31E-03 &    4.36E-03 &   4.70E-03 &   4.93E-03 \\
 \cline{2-7}
           &    \#speedup &       \bf 7.89 &      \bf 9.80 &    \bf  11.99 &    \bf  14.04 &   \bf   16.94 \\
\hline
\end{tabular}
\end{scriptsize}
\end{table}

Note that it takes only 46.27 seconds to calculate $\bA^{\top}\bA$. In other words, the consumed time per signal is only 0.23 seconds. %
If there are 200,000 signals, then the computational time per signal will be $2.3 \times 10^{-4}$ seconds, %
which is negligible.

\subsection{Many-face Recognition by BMPL}

We apply BMPL for many-face recognition tasks by solving problem (\ref{eq:batch}). We adopt {L2}~\cite{shi2011}, {L2-L2}~\cite{Lei2011} and
{BOMP}~\cite{Rubinstein2008} as the baseline methods.  Besides, the PGH method is adopted for the comparison, since it has shown better efficiency than other $\ell_1$-norm methods~\cite{linxiao,Xiao2013PGH_J}.   We follow
the experimental settings in~\cite{Wright2009} for the comparison.  which is negligible. We  set $\varrho = 10$ for {BMPL} and
$k=200$ for {BOMP} for all experiments. Furthermore, considering
that there may be some images that cannot be sparse-represented by
the training images, we constrain $k \leq 600$.

\begin{table*}[htp]
\center
\begin{scriptsize}
\caption{Prediction Accuracy on Two Face
Databases}\label{table:acc_YaleB}
\begin{tabular}{|c|c|c|c|c|c|c|c||c|c|c|c|c|}
\hline
           &                                            \multicolumn{ 7}{|c||}{Extended  {YaleB} Database} &                             \multicolumn{ 5}{|c|}{AR Database} \\
\hline
      $\rho_{d}$      &       1 &     1/2 &     1/3 &     1/4 &     1/5 &     1/6 &    1/7 &      1 &    3/4 &   2/3 &    1/2 &    1/3 \\
\hline
        L2 &     0.9876 &     0.9868 &     0.9831 &     0.9792 &     0.9371 &     0.9561 &     0.9621 &     0.9466 &     0.9301 &     0.9108 &    \textbf{0.7323}&     0.9638 \\
\hline
     L2-L2 &     0.9898 &     0.9859 &     0.9827 &     0.9818 &     0.9783 &     0.9730 &     0.9723 &     0.9524 &     0.9504 &     0.9532 &     0.9574 &     0.9692 \\
\hline
       PGH &     0.9897 &     0.9843 &     0.9826 &     0.9846 &     0.9815 &     0.9760 &     0.9658 &     0.9657 &     0.9650 &     0.9715 &     0.9679 &     0.9656 \\
\hline
      BOMP &     0.9904 &     0.9897 &     0.9861 &     0.9844 &     0.9786 &     0.9799 &     0.9734 &     0.9742 &     0.9744 &     0.9738 &     0.9738 &     0.9619 \\
\hline
      BMPL &     0.9911 &     0.9892 &     0.9873 &     0.9849 &     0.9817 &     0.9787 &     0.9761 &     0.9739 &     0.9757 &     0.9715 &     0.9723 &     0.9672 \\
\hline
   Wilcoxon &          0 &          0 &          0 &          0 &          1 &          1 &          1 &          1 &          1 &          1 &          1 &          0 \\
\hline
\end{tabular}
\end{scriptsize}
\end{table*}

The \emph{{Extended YaleB}} and
\emph{AR} databases are used for the comparison. The \emph{{Extended YaleB}} database
consists of 2,414 frontal face images of 38
subjects~\cite{shi2011,Gao2013}. They are captured under various
lighting conditions and cropped and normalized to $192\times168$
pixels. In our experiment, we take 62 images per person, resulting
in 2,356 images in total. The \emph{AR} database consists of over
2,600 frontal images of 100
individuals~\cite{Martinez1999,Wright2009,Gao2013}. Each image is
normalized to $80\times60$ pixels. {Computing $\bA^{\top}\bA$ with all images
of \emph{{Extended YaleB}} and \emph{AR} takes 5.74 seconds and 1.10
seconds, respectively. In other words, the time spent on
$\bA^{\top}\bA$ is negligible.

We consider two experimental settings: 1)
\emph{Many-face recognition with different number of pixels}; and 2) \emph{Many-face
recognition with different number of training samples}.

\begin{table*}[htp]
\center \caption{Total Time Spent on Two Face Databases (in
seconds), \#speedup denotes the times of speedup of BMPL over
PGH}\label{table:time_YaleB}
\begin{scriptsize}
\begin{tabular}{|c|c|c|c|c|c|c|c||c|c|c|c|c|}
\hline
           &                                           \multicolumn{ 7}{|c||}{Extended  {YaleB} Database} &                             \multicolumn{ 5}{|c|}{AR Database} \\
\hline
      $\rho_{d}$   &       1 &     1/2 &     1/3 &    1/4 &     1/5 &    1/6 &    1/7 &      1 &    3/4 &    2/3 &    1/2 &    1/3 \\
\hline
        L2 &      71.33 &      24.91 &       6.29 &       3.51 &       2.42 &       1.14 &       0.72 &      13.34 &       4.39 &       3.16 &       3.28 &       2.19 \\
\hline
     L2-L2 &      11.36 &       6.85 &       4.13 &       2.40 &       2.32 &       2.22 &       1.69 &       3.75 &       3.04 &       3.10 &       2.58 &       1.99 \\
\hline
       PGH &    5559.53 &    4863.18 &    2195.03 &    1383.28 &     822.11 &     627.95 &     383.86 &    5229.75 &    2812.96 &    2178.91 &    1324.59 &     557.65 \\
\hline
      BOMP &     139.69 &      99.88 &      98.05 &      89.83 &      89.95 &      90.41 &      87.60 &     108.52 &      98.84 &      98.60 &      97.25 &      95.58 \\
\hline
      BMPL &      39.72 &      17.05 &      12.94 &       7.86 &       7.62 &       6.53 &       6.19 &      14.29 &      10.87 &      10.20 &       7.14 &       4.57 \\
\hline
      \#speedup &      140.0 &     \textbf{ 283.6} &      169.6 &      176.0 &       107.9 &       96.2 &       62.0 &      \textbf{366.0} &       258.8 &      213.6 &       185.5 &        122.0 \\
 \hline
\end{tabular}
\end{scriptsize}
\end{table*}

\begin{table*}[htp]
\center \caption{Average Sparsity on Two Face Databases
}\label{table:spar_YaleB}
\begin{scriptsize}
\center
\begin{tabular}{|c|c|c|c|c|c|c|c||c|c|c|c|c|}
\hline
           &                                           \multicolumn{ 7}{|c||}{Extended YaleB Database} &                             \multicolumn{ 5}{|c|}{AR Database} \\
\hline
     $\rho_{d}$   &       1 &     1/2 &     1/3 &    1/4 &     1/5 &    1/6 &    1/7 &      1 &    3/4 &    2/3 &    1/2 &    1/3 \\
\hline
      BOMP &        200 &        200 &        200 &        200 &        200 &        200 &        200 &        200 &        200 &        200 &        200 &        200 \\
\hline
       PGH &        164 &        165 &        165 &        162 &        156 &        158 &        163 &        133 &        130 &        127 &        135 &        124 \\
\hline
       BMPL &        167 &        165 &        160 &        155 &        155 &        149 &        143 &        189 &        190 &        188 &        194 &        201 \\
\hline
\end{tabular}
\end{scriptsize}
\end{table*}

\subsubsection{Many-face Recognition with Different Number of Pixels}
In this experiment, we
down-sample the images at a sampling rate $\rho_{d}$, where
$\rho_{d}$ is chosen from $\{1,  1/2, 1/3,    1/4, 1/5, 1/6, 1/7\}$
for \emph{YaleB} images, and $\{1, 3/4,    2/3, 1/2, 1/3\}$ for
\emph{AR} images. Accordingly, the dimension of each new image vector will
be $\rho_{d}^2$ of the original image vector.
Following~\cite{shi2011}, we randomly choose half of the  images of each
person as the training set, and the remaining images as the testing set.  The prediction
accuracies on the \emph{YaleB} and \emph{AR} images are shown in Table
\ref{table:acc_YaleB}. To measure the difference between results, the
\emph{Wilcoxon} test with $5\%$ significance is conducted between
{BMPL} and  the winner of {L2} and {L2-L2}, and  1 indicates the
significant difference.

From Table \ref{table:acc_YaleB}, on the \emph{YaleB} database, {BMPL}
shows significantly better accuracy than {L2} and {L2-L2} methods
under  $\rho_{d} = 1/5, 1/6 $ and $1/7$, and comparable or slightly
better performance under other down-sampling rates. On  the \emph{AR}
database, {BMPL} performs significantly better than {L2} and {L2-L2} methods
under $\rho_{d} = 1, 3/4 $ and $2/3$. {BMPL} in particular
shows much more stable performance than the {L2} and {L2-L2} methods. In particular, on the \emph{AR} database, {L2} only achieves
$\textbf{73.23}\%$ prediction accuracy at a down-sampling rate
$\rho_{d}=1/2$, which may be caused by the unstable pseudo inverse on the
ill-conditioned matrix~\cite{shi2011}. As a regularized {L2} method,
{L2-L2} method shows more stable
performance than {L2}. However, it is still worse than {BMPL}.

We report the total time spent by various methods in Table
\ref{table:time_YaleB}. {PGH}, the state-of-the-art $\ell_1$-solver,
needs several hours to predict all testing images on the \emph{AR} database
with $\rho_d = 1$, which is unbearable for many real-world
applications. On the contrary, BMPL completes the prediction in 20
seconds only, which is 366 times faster than {PGH}. BMPL is
also 3-10 times faster than {BOMP}. Lastly, BMPL
achieves comparable efficiency to {L2-L2} and {L2}.

A remaining question is: \emph{does the sparsity help to
improve  recognition performance}?  We list the average sparsity of {BMPL}, {PGH}, and {BOMP} in Table \ref{table:spar_YaleB}. Note that the solutions obtained by {L2} and {L2-L2} methods are not sparse.  From Table \ref{table:acc_YaleB}, {BMPL},
{PGH}, and {BOMP} show comparable or significantly better
recognition rates than {L2} and {L2-L2} methods on the \emph{YaleB}
database. In addition, {BMPL} outperforms {L2} and {L2-L2} methods on
{\emph{AR}} database with enough pixels. Therefore, sparsity indeed \textbf{{helps} to improve}  recognition rates.

\subsubsection{Face Recognition with Different Number of Training Samples}

Let $\rho_t$ be the ratio of the number of training
images over the total number of images. In this experiment, we vary $\rho_t \in \{0.55,
0.60,0.65, 0.7, 0.75,0.8\}$ to change
the number of training images.  The prediction accuracy and prediction
time w.r.t. $\rho_t$ are shown in Tables
\ref{table_acc_Yele_train} and \ref{table_time_Yele_train},
respectively.

In general, with
more training images, the matrix $\bA^{\top}\bA$ becomes
more ill-conditioned.
 From Table \ref{table_acc_Yele_train}, {BMPL} {performs}
significantly better than {L2} and {L2-L2} when $\rho_t \geq  0.60$.
In other words, {BMPL} achieves more stable performance when
$\bA^{\top}\bA$ becomes more ill-conditioned. Finally,  from Table
\ref{table_time_Yele_train}, {BMPL} shows
comparable efficiency to {L2} and {L2-L2} methods.


\begin{table} \center \caption{Prediction Accuracy on \emph{YaleB}
 with Different Number of Training
Images}\label{table_acc_Yele_train}
\begin{scriptsize}
\begin{tabular}{|c|c|c|c|c|c|c|}
\hline
   $\rho_{t}$      &       0.55 &       0.60 &       0.65 &       0.70 &       0.75 &       0.80 \\
\hline
        L2 &     0.6352 &     0.9350 &     0.9330 &     0.9684 &     0.9764 &     0.9815 \\
\hline
     L2-L2 &     0.9814 &     0.9814 &     0.9823 &     0.9827 &     0.9843 &     0.9872 \\
\hline
       BMPL &     0.9848 &     0.9887 &     0.9887 &     0.9908 &     0.9911 &     0.9925 \\
\hline
   Wilcoxon &          0 &          1 &          1 &          1 &          1 &          1 \\
\hline
\end{tabular}
\end{scriptsize}
\end{table}
\begin{table}[htp]
\center \caption{Total Time Spent on \emph{YaleB} with Different
Number of Training Images (in seconds)}\label{table_time_Yele_train}
\begin{scriptsize}
\begin{tabular}{|c|c|c|c|c|c|c|}
\hline
     $\rho_{t}$       &       0.55 &       0.60 &       0.65 &       0.70 &       0.75 &       0.80 \\
\hline
        L2 &       2.48 &       2.95 &       3.02 &       3.16 &       3.56 &       6.02 \\
\hline
     L2-L2 &       2.20 &       2.51 &       3.50 &       3.94 &       3.21 &       6.06 \\
\hline
       BMPL &      10.65 &       6.11 &       5.71 &       4.93 &       4.23 &       2.85 \\
\hline
\end{tabular}

\end{scriptsize}
\end{table}

\section{Conclusions}\label{sec:con}
In this paper, we have proposed a subspace  search to  further improve the
performance of MPL, and a batch-mode MPL has been developed to
vastly speed up SR with many signals. Comprehensive experiments demonstrate the superb efficiency of the proposed (S)MPL methods.
In general, (S)MPL are tens times faster than state-of-the-art
$\ell_1$-norm methods. The recovery time of the SMPL
method over a \emph{Big Dictionary} with {one million atoms is less
than 50 seconds}. We apply {BMPL} to batch face recognition tasks.
The experimental results show that BMPL achieves significantly
better recognition rates than {L2} and {L2-L2} with comparable
computational cost. Notably, BMPL is up to 20 times faster than
the batch-mode OMP~\cite{Rubinstein2008} and 400 times faster than
the $\ell_1$-norm methods  considered to be state-of-the-art.

\section*{Acknowledgement}
The authors would like to thank the anonymous reviewers
for their insightful comments and suggestions which have
greatly improved the paper.
This research was partially supported by the Australian Research Council
Future Fellowship FT130100746, Australian Research Council grants DE120101161, and DP140102270.
\ifCLASSOPTIONcaptionsoff
  \newpage
\fi

\end{document}